\documentclass[12pt,journal,compsoc]{IEEEtran}
\ifCLASSOPTIONcompsoc
   \usepackage[nocompress]{cite}
\else

   \usepackage{cite}
\fi
\usepackage[pdftex]{graphicx}
\usepackage[cmex10]{amsmath}
\usepackage{amsthm}
\usepackage{lineno,hyperref}
\usepackage{bm}
\usepackage{algorithm}
\usepackage{algpseudocode}
\usepackage{lipsum}
\usepackage{varwidth}

\newcommand{\quotes}[1]{``#1''}
\usepackage{array}
\usepackage[font=footnotesize]{subfig}
\usepackage{fixltx2e}
\usepackage{color}
\usepackage{amsfonts}
\usepackage{rotating}
\makeatletter
\newsavebox\myboxA
\newsavebox\myboxB
\newlength\mylenA

\newcommand*\xoverline[2][0.75]{%
    \sbox{\myboxA}{$\m@th#2$}%
    \setbox\myboxB\null% Phantom box
    \ht\myboxB=\ht\myboxA%
    \dp\myboxB=\dp\myboxA%
    \wd\myboxB=#1\wd\myboxA% Scale phantom
    \sbox\myboxB{$\m@th\overline{\copy\myboxB}$}%  Overlined phantom
    \setlength\mylenA{\the\wd\myboxA}%   calc width diff
    \addtolength\mylenA{-\the\wd\myboxB}%
    \ifdim\wd\myboxB<\wd\myboxA%
       \rlap{\hskip 0.5\mylenA\usebox\myboxB}{\usebox\myboxA}%
    \else
        \hskip -0.5\mylenA\rlap{\usebox\myboxA}{\hskip 0.5\mylenA\usebox\myboxB}%
    \fi}
\makeatother

\begin{document}

\title{Causality on Longitudinal Data:\\ Stable Specification Search in Constrained Structural Equation Modeling}

\author{Ridho~Rahmadi, Perry~Groot, Marieke~HC~van~Rijn, Jan~AJG~van~den~Brand, Marianne~Heins, Hans~Knoop, Tom~Heskes,\\ the Alzheimer's~Disease~Neuroimaging~Initiative$^*$,\\ the~MASTERPLAN~Study~Group, the OPTIMISTIC~consortium$^{**}$
\IEEEcompsocitemizethanks{\IEEEcompsocthanksitem R. Rahmadi is with the Department
of Informatics, Universitas Islam Indonesia and Institute for Computing and Information Sciences, Radboud University Nijmegen, the Netherlands. E-mail: r.rahmadi@cs.ru.nl % \protect \\
\IEEEcompsocthanksitem P. Groot and T. Heskes are with Institute for Computing and Information Sciences, Radboud University Nijmegen, the Netherlands.
\IEEEcompsocthanksitem M. van Rijn and J. van~den~Brand are with Department of Nephrology, Radboud University Medical Center, Nijmegen, The Netherlands.
\IEEEcompsocthanksitem H. Knoop is with the Department of Medical Psychology, Academic Medical Center, University of Amsterdam, The Netherlands.
\IEEEcompsocthanksitem $^{**}$The members of OPTIMISTIC consortium are described in the acknowledgment.
\IEEEcompsocthanksitem $^{*}$One of the data sets used in preparation of this article were obtained from the Alzheimer's Disease Neuroimaging Initiative (ADNI) database (adni.loni.usc.edu). As such, the investigators within the ADNI contributed to the design and implementation of ADNI and/or provided data but did not participate in analysis or writing of this report. A complete listing of ADNI investigators can be found at: \url{http://adni.loni.usc.edu/wp-content/uploads/how_to_apply/ADNI_Acknowledgement_List.pdf}
\IEEEcompsocthanksitem M. Heins is with the Netherlands Institute for Health Services Research, The Netherlands.
}}

\IEEEtitleabstractindextext{
\begin{abstract}
A typical problem in causal modeling is the instability of model structure learning, i.e., small changes in finite data can result in completely different optimal models.
The present work introduces a novel causal modeling algorithm for longitudinal data, that is robust for finite samples based on recent advances in stability selection using subsampling and selection algorithms. Our approach uses exploratory search but allows incorporation of prior knowledge, e.g., the absence of a particular causal relationship between two specific variables.
We represent causal relationships using structural equation models. Models are scored along two objectives: the model fit and the model complexity. Since both objectives are often conflicting we apply a multi-objective evolutionary algorithm to search for Pareto optimal models. To handle the instability of small finite data samples, we repeatedly subsample the data and select those substructures (from the optimal models) that are both stable and parsimonious. These substructures can be visualized through a causal graph.
Our more exploratory approach achieves at least comparable performance as, but often a significant improvement over state-of-the-art alternative approaches on a simulated data set with a known ground truth. We also present the results of our method on three real-world longitudinal data sets on chronic fatigue syndrome, Alzheimer disease, and chronic kidney disease. The findings obtained with our approach are generally in line with results from more hypothesis-driven analyses in earlier studies and suggest some novel relationships that deserve further research.
\end{abstract}

}

\maketitle
\IEEEdisplaynontitleabstractindextext
\IEEEpeerreviewmaketitle

\section{Introduction}Causal modeling, an essential problem in many disciplines\cite{daniel2012using,hoover2008causality,abu2003government,taguri2015causal,pearl1995causal,detilleux2016bayesian}  attempts to model the mechanisms by which variables relate and to understand the changes on the model if the mechanisms were manipulated\cite{spirtes2010introduction}.
In the medical domain,
revealing causal relationships may lead to improvement of clinical practice, for example, the development of
treatment and medication.
Slowly but steadily, causal discovery methods find their way into the medical literature, providing novel insights through exploratory analyses\cite{la2014causal,sokolova2014causal,cooper2015center}. Moreover, data in the medical domain is often collected through longitudinal studies. Unlike in a cross-sectional design, where all measurements are obtained at a single occasion, the data in a longitudinal design consist of repeated measurements on subjects through time. Longitudinal data make it possible to capture change within subjects over time and thus gives some advantage to causal modeling in terms of providing more knowledge to establish causal relationships\cite{frees2004longitudinal}. As emphasized in Fitmaurice et al.,\cite{fitzmaurice2012applied} there is much natural heterogeneity among subjects in terms of how diseases progress that can be explained by the longitudinal study design. Another advantage is that in order to obtain a similar level of statistical power as in cross-sectional studies, fewer subjects in longitudinal studies are required\cite{hedeker2006longitudinal}.

To date, a number of causal modeling methods have been developed for longitudinal (or time series) data. Some of the methods are based on a Vector Autoregressive (VAR) and/or Structural Equation Model (SEM) framework which assumes a linear system and independent Gaussian noise\cite{swanson1997impulse,bessler2002money,demiralp2003searching,moneta2008graphical,kim2007unified}.
Some other methods, interestingly, take advantage of nonlinearity \cite{moneta2011causal,peters2013causal,chu2008search}, or non-Gaussian noise \cite{peters2013causal,hyvarinen2008causal}, to gain even more causal information. Most of the aforementioned methods conduct the estimation of the causal structures in somewhat similar ways. For example \cite{bessler2002money,demiralp2003searching,moneta2008graphical,peters2013causal,hyvarinen2008causal} use the (partial correlations of the) VAR residuals to either test independence or as input to a causal search algorithm, e.g., LiNGAM (Linear Non-Gaussian Acyclic Model)\cite{shimizu2006linear}, PC (\quotes{P} stands for Peter, and \quotes{C} for Clark, the authors)\cite{spirtes2000causation}. In general these causal search algorithms are solely based on a \emph{single run} of model learning which is notoriously instable: small changes in finite data samples can lead to entirely different inferred structures. This implies that, some approaches might not be robust enough to correctly estimate causal models from various data, especially when the data set is noisy or has small sample size.

In the present paper, we introduce a robust causal modeling algorithm for longitudinal data that is designed to resolve the instability inherent to structure learning. We refer to our method as S3L, a shorthand for Stable Specification Search for Longitudinal data. It extends our previous method\cite{Rahmadi2016}, here referred to as S3C, which is designed for cross-sectional data. S3L is a \emph{general framework} which subsamples the original data into many subsets, and for each subset S3L heuristically searches for \emph{Pareto optimal} models using a multi-objective optimization approach. Among the optimal models, S3L observes the so-called \emph{relevant} causal structures which represent both stable and parsimonious model structures. These steps constitute the structure estimation of S3L which is fundamentally different from the aforementioned approaches that mostly use a single run for model estimation. For completeness, detail about S3C/L is described in Section~\ref{sec:background}. Moreover, in the default setting S3L assumes some underlying contexts: iid samples for each time slice (lag), linear system, additive independent Gaussian noise, causal sufficiency (no latent variables), stationary (time-invariant causal relationships), and fairly uniform time intervals between time slices.

The main contributions of S3L are:
\begin{itemize}
  \item The causal structure estimation of S3L is conducted through multi-objective optimization and stability selection\cite{meinshausen2010stability} over optimal models, to optimize both the stability and the parsimony of the model structures.
  \item S3C/L is a general framework which allows for other causal methods with all of their corresponding assumptions, e.g., nonlinearity, non-Gaussianity, to be plugged in as model representation and estimation. The multi-objective search and the stability selection part are independent of any mentioned assumptions.
  \item In the default model representation, S3L adopts the idea of the \quotes{rolling} model from~\cite{friedman1998learning} to transform a longitudinal SEM model with an arbitrary number of time slices into two parts: a baseline model and a transition model. The baseline model captures the causal relationships at baseline observations, when subjects enter the study. The transition model consists of two time slices, which essentially represent the possible causal relationships within and across time slices. We also describe how to reshape the longitudinal data correspondingly, so as to match the transformed longitudinal model which then can easily be scored using standard SEM software.
  \item We provide standardized causal effects which are computed from IDA estimates\cite{maathuis2009estimating}.
  \item We carry out experiments on three different real-world data of (a) patients with chronic fatigue syndrome (CFS), (b) patients with Alzheimer disease (AD), and (c) patients with chronic kidney disease (CKD).
\end{itemize}

Some relevant methods, have attempted to make use of common structures to infer causal models. Causal Stability Ranking (CStaR)\cite{stekhoven2012causal}, originally designed for gene expression data, tries to find stable rankings of genes (covariates) based on their total causal effect on a specific phenotype (response), using a subsampling procedure similar to stability selection and IDA to estimate causal effects. As CStaR only focuses on relationships from all covariates to a single specific response, it seems to be difficult to generalize it to other domains where any possible causal relationship may be of interest. Moreover, another approach called Group Iterative Multiple Model Estimation (GIMME), originally developed for functional Magnetic Resonance Imaging (fMRI) data and essentially an extension of \emph{extended unified} SEM (combination of VAR and SEM)\cite{gates2011extended}, aims to combine the group-level causal structures with the individual-level structures, resulting in a causal model for each individual which contains common structures to the group.
Such subject-specific estimation may be feasible given relatively long time series (as in resting state fMRI), but likely too challenging for the typical longitudinal data in clinical studies with a limited number of time slices per subject. Still in the domain of fMRI, there is a method called Independent Multiple-sample Greedy Equivalence Search (IMaGES)\cite{ramsey2010six}. The method is a modification of GES (described in the following paragraph), and designed to handle unexpected statistical dependencies in combined data. Since IMaGES was developed mainly for combining results of multiple data sets, we do not consider it further.

Having both the transformed longitudinal model and the reshaped data, we can run other alternative approaches which are designed for cross-sectional data and conduct comprehensive comparisons. Here, for evaluation of S3L, we generate simulated data and compare with some advanced constrained-based approaches such as PC-stable\cite{colombo2014order}, Conservative PC (CPC)\cite{ramsey2012adjacency}, CPC-stable\cite{colombo2014order,ramsey2012adjacency}, and PC-Max\cite{ramsey2016improving}. All of these methods are extensions of the PC algorithm which in principle consists of two stages. The first stage uses conditional independence tests to obtain the skeleton (undirected edges) of the model, and the second stage orients the skeleton based on some rules, resulting in an \emph{essential graph} or \emph{markov equivalence class} model (described in Section \ref{sec:stableSpecSearch}, for more detail see \cite{chickering2002learning}). We also compare with an advanced score-based algorithm called Fast Greedy Equivalent Search (FGES) \cite{ramseymillion}. It is an extension of GES which in general starts with an empty (or sparse) model, and iteratively adds an edge (forward phase) which mostly increases the score until no more edge can be added. Then GES iteratively prunes an edge (backward phase) which does not decrease/improve the score until no more edge can be excluded.

The rest of this paper is organized as follows. All methods used in our approach are presented in  Section~\ref{sec:background}. The results and the corresponding discussions are presented in Section~\ref{sec:experimental}. Finally, conclusions and future work are presented in Section~\ref{sec:conclusion}.

\section{Methods}
\label{sec:background}
\subsection{Stable Specification Search for Cross-Sectional data}
\label{sec:stableSpecSearch}
In \cite{Rahmadi2016} we introduced our previous work, S3C, which searches over structures represented by \sloppy SEMs. In SEMs, refining models to improve the model quality is called \emph{specification search}. Generally S3C adopts the concept of stability selection\cite{meinshausen2010stability} in order to enhance the robustness of structure learning by considering a whole range of model complexities. Originally, in stability selection, this is realized by varying a continuous regularization parameter. Here, we explicitly consider different discrete model complexities. However, to find the optimal model structure for each model complexity is a hard optimization problem. Therefore, we rephrase stability selection as a multi-objective optimization problem, so that we can jointly run over the whole range of model complexities and find the corresponding optimal structures for each model complexity.

In more detail, S3C can be divided into two phases. The first phase is \emph{search}, performing exploratory search over Structural Equation Models (SEMs) using a multi-objective evolutionary algorithm called Non-dominated Sorting Genetic Algorithm II \mbox{(NSGA-{II}})\cite{deb2002fast}. \mbox{NSGA-{II}} is an iterative procedure which adopts the idea of evolution. It starts with random models and in every \emph{generation} (iteration), attempts to improve the quality of the models by manipulating (refining) good models (parents) to make new models (offsprings). The quality of the models is characterized by scoring that is based on two conflicting objectives: model fit with respect to the data and model complexity. The model manipulations are realized by using two genetic operators: \emph{crossover} that combines the structures of parents and \emph{mutation} that flips the structures of models. Moreover, the composition of model population in the next generation is determined by \emph{selection} strategy. One of the key features of NSGA-{II} is that in every iteration, it sorts models based on the concept of \emph{domination}, yielding \emph{fronts} or sets of models such that models in front $l$ dominate those in front $l+1$. The domination concept states that model $m_1$ is said to dominate model $m_2$ if and only if model $m_1$ is no worse than $m_2$ in all objectives and the model $m_1$ is strictly better than $m_2$ in at least one objective. The first front of the last generation is called the \emph{Pareto optimal} set, giving optimal models for the whole range of model complexities. Details of the NSGA-{II} algorithm are described in Deb et al\cite{deb2002fast}.

Based on the idea of stability selection\cite{meinshausen2010stability}, S3C subsamples $N$ subsets from the data $D$ with size $\lfloor |D|/2 \rfloor$ without replacement, and for each subset, the search phase above is applied, giving sets of Pareto optimal models.
After that, all Pareto optimal models are transformed into their corresponding Markov equivalence classes which can be represented by \emph{Completed Partially Directed Acyclic Graphs} (CPDAGs)\cite{chickering2002learning}. Since all DAGs that are a member of the same Markov equivalence class represent the same probability distribution, they are indistinguishable based on the observational data alone. In SEMs, these models are called covariance equivalent\cite{pearl2000causality} and return the same scores.
From these CPDAGs we compute the \emph{edge} and \emph{causal path stability} graphs (see Figure~\ref{stabGraphCFS} for an example)
by grouping them according to model complexity and computing their \emph{selection probability}, i.e., the number of occurrences divided by the total number of models for a certain level of model complexity.
The edge stability considers any edge between a pair of variables (i.e., $A\to B$, $B\to A$, or $A-B$) and the causal path stability considers directed path, e.g., $A\to B$ of any length.
Stability selection is then performed by specifying two thresholds, $\pi_{\mathrm{sel}}$ (boundary of selection probability) and $\pi_{\mathrm{bic}}$ (boundary of complexity). For example, setting $\pi_{\mathrm{sel}}=0.6$ means that all causal relationships with edge stability or causal path stability greater than or equal to this threshold are considered \emph{stable}. The second threshold $\pi_{\mathrm{bic}}$ is used to control overfitting.
For every model complexity $j$, we compute the Bayesian Information Criterion (BIC) score for each model in $j$ based on the data subset to which the model is fitted. We then compute $\xoverline[0.9]{BIC}_j$ the average of BIC scores in model complexity $j$. We set $\pi_{\mathrm{bic}}$ to the minimum $\xoverline[0.9]{BIC}_j$.
All causal relationships with an edge stability or a causal path stability that is smaller than or equal to $\pi_{\mathrm{bic}}$ (e.g., \mbox{$\pi_{\mathrm{bic}}=27$} in Figure~\ref{edgeStabCFSTransition}) are considered \emph{parsimonious}.
Hence, the causal relationships greater than or equal to $\pi_{\mathrm{sel}}$ and smaller than or equal to $\pi_{\mathrm{bic}}$ are considered both stable and parsimonious and called \emph{relevant} from which we can derive a causal model. In addition, we call the region with which the relevant structures intersect as relevant region.

The second phase concerns \emph{visualization}, combining the stability graphs into a graph with nodes and edges. This is done by adding the relevant edges and orienting them using \emph{prior knowledge} described in Section~\ref{sec:constrainedSEM}) and the relevant causal paths. More specifically, we
first connect the nodes following the relevant edges. Then we orient these edges based
on the prior knowledge. And finally, we orient the rest of the edges following the relevant causal paths.
The resulting graph consists of directed edges which represent causal relationship and possibly with additional undirected edges which represent strong association but for which the direction is unclear from the data.
Furthermore, following Meinshausen and
B\"{u}hlmann\cite{meinshausen2010stability}, for each edge in the graph we take the highest selection probability it has across different model complexities in the relevant region of the edge stability graph as a measure of \emph{reliability} and annotate the corresponding edge with this reliability score.
The reliability score indicates the confidence of a particular relevant structure. The higher the score, the more we can expect that the relevant structure is not falsely selected\cite{meinshausen2010stability}.
In addition each directed edge is annotated with a standardized causal effect estimate which is explained in Section \ref{causalEffect}. The stability graphs are considered to be the main outcome of our approach where the visualization eases interpretation.

\subsection{Stable specification search for longitudinal data}
\begin{figure*}[!htbp]
\centering
\includegraphics[width=0.75\textwidth]{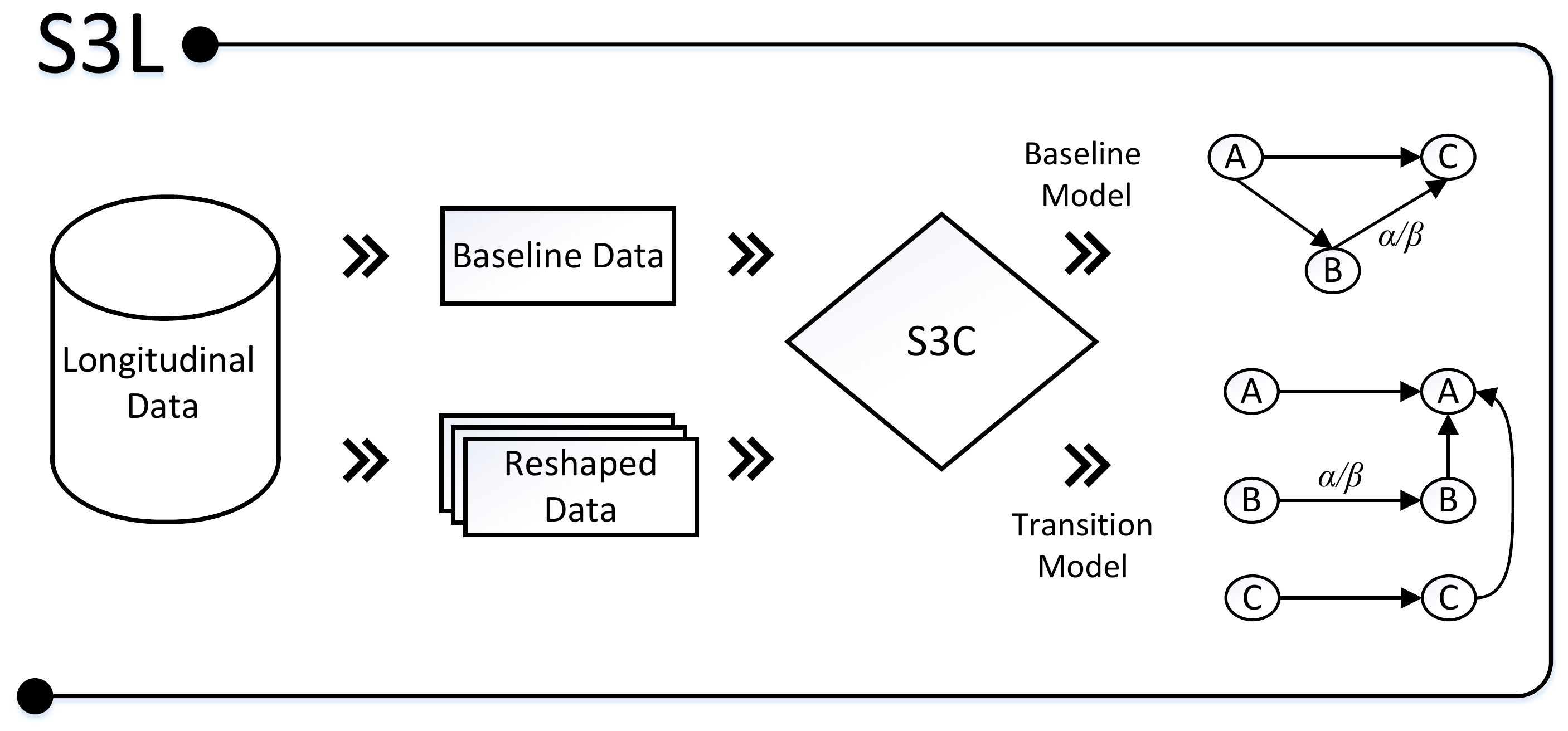}
\caption{Given a longitudinal data set, S3L uses the baseline observations to infer a baseline model, and reshapes the whole data set to infer a transition model. Both baseline and transition model are annotated with a reliability score $\alpha$ and a standardized causal effect $\beta$.}
\label{fig_S3L}
\end{figure*}

Stable Specification Search for Longitudinal data (S3L) is an extension of S3C. In principle, as illustrated in Figure~\ref{fig_S3L}, S3L applies S3C on transformed longitudinal models, called baseline and transition models (explained in Section~\ref{sec:longitudinalModel}). Furthermore, in order to see to which extent a covariate would cause a response, S3L provides standardized total causal effect estimates which are intrinsically computed from estimates from IDA\cite{maathuis2009estimating} (described in Section~\ref{causalEffect}). In the following subsections, we first describe how we transform a longitudinal model and reshape the data accordingly, and then we discuss
the implication of allowing prior knowledge in our S3C structure learning.

\subsubsection{Longitudinal model and data reshaping}
\label{sec:longitudinalModel}
\begin{figure*}[!tbp]%[!t] %p
\centering
\mbox{\subfloat[]
{
\includegraphics[page=1,width=0.115\textwidth]{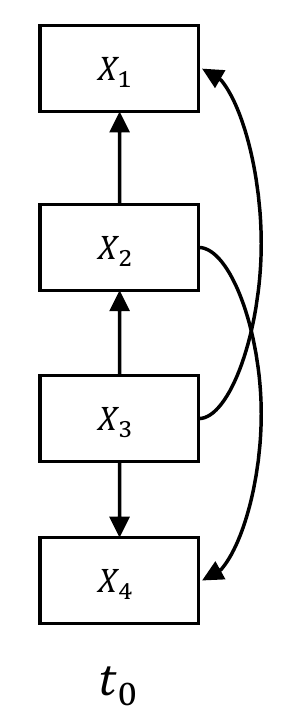} %0.18
\label{prior_model}
}}
\qquad
\mbox{\subfloat[]
{
\includegraphics[page=2,width=0.2\textwidth]{rolled_network} %0.18
\label{transition_model}
}}
\qquad
\mbox{\subfloat[]
{
\includegraphics[page=3,width=0.32\textwidth]{rolled_network} %0.26
\label{unrolled_model}
}}
\caption{(a) The baseline model which is used to capture causal relationships at the initial time slice, e.g., before medical treatment. (b) The transition model which is used to represent causal relationships within and between time slices, e.g., during medical treatment. (c) The corresponding \quotes{unrolled} longitudinal model.}
\end{figure*}

Based on the idea of a \quotes{rolling} network in~\cite{friedman1998learning} we transform a longitudinal SEM with an arbitrary number of time slices (e.g., Figure~\ref{unrolled_model}) into two parts: a baseline model (Figure~\ref{prior_model}) and a transition model (Figure~\ref{transition_model}). In the original paper, the authors treat these models as probabilistic networks, here we treat them purely as SEMs. The baseline model essentially represents the causal relationships between variables that may happen at the initial time slice $t_0$, for instance, causal relationships that occur before a medical treatment started. Moreover, the baseline model may also represent relationships of the unobserved process before $t_0$\cite{friedman1998learning}. The transition model constitutes the causal relationships between variables across time slices $t_{i-1}$ and $t_i$, and between variables within time slice $t_i$ for $i>0$, for example, causal relationships that represent interactions during a medical treatment. In S3L, the structure estimations will be conducted on the baseline and transition model separately.

From the transition model we distinguish two kinds of causal relationships, namely \emph{intra-slice} causal relationship (e.g., solid arcs in Figure~\ref{transition_model}), and \emph{inter-slice} causal relationship (e.g., dashed arcs in Figure~\ref{transition_model}). The intra-slice causal relationship represents relationships \emph{within} time slice  $t_i$. Accordingly the inter-slice causal relationship represents relationships \emph{between} time slices $t_{i-1}$ and $t_i$. We assume that the inter-slice causal relationships are independent of $t$ (stationary).
We also assume that the time intervals between time slices are fairly uniform. In addition, the transition model implies two more constraints (explained in Section~\ref{sec:constrainedSEM}): there is no intra-slice causal relationship allowed in time slice $t_{i-1}$ and the inter-slice causal relationships always go forward in time, i.e., from time slice $t_{i-1}$ to time slice $t_i$.

\begin{figure*}[!htbp]
\centering
\includegraphics[width=0.7\textwidth]{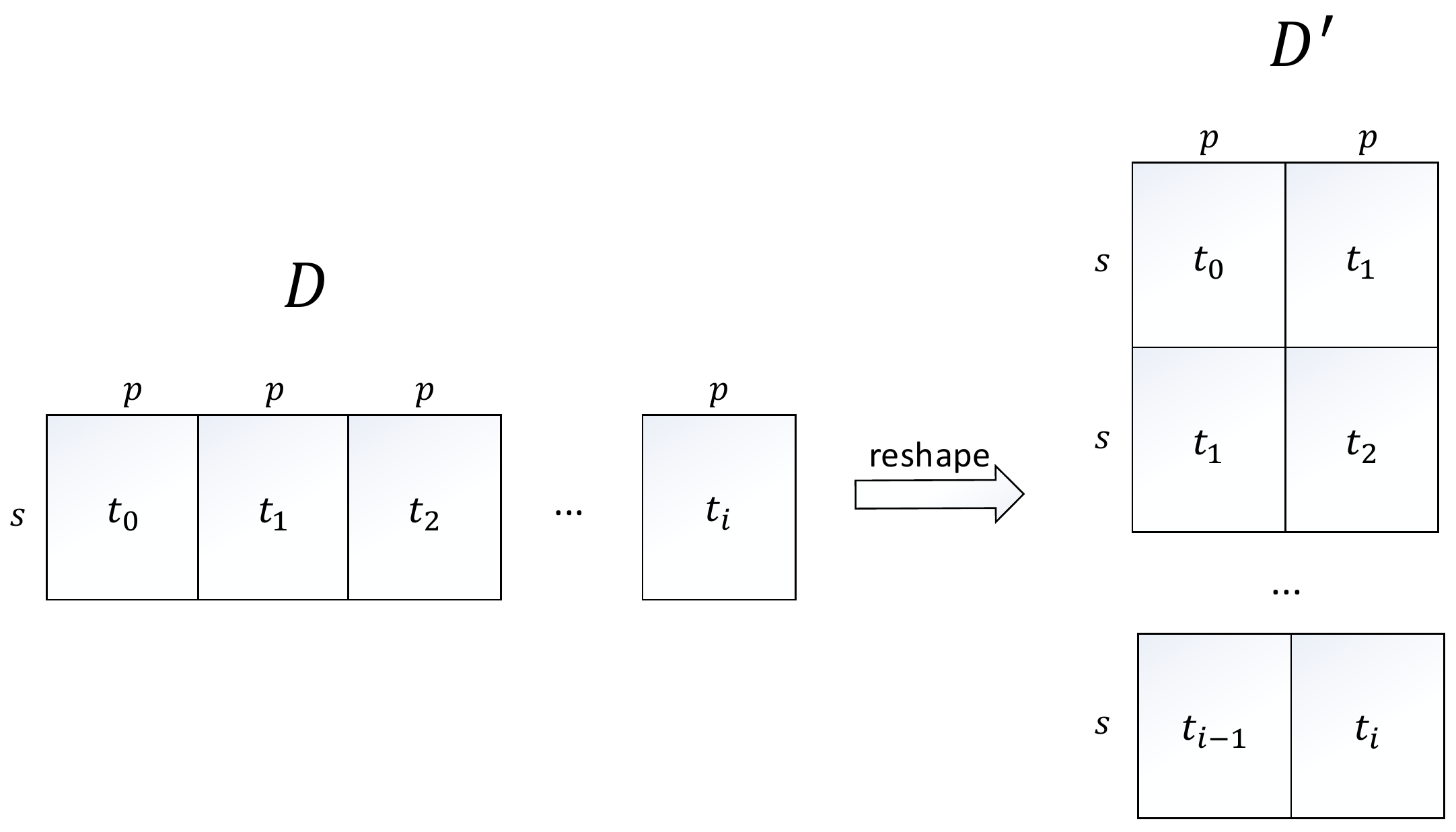}
\caption{$D$ is a matrix representing the original data shape which consists of $s$ instances, $p$ variables, and $i$ time slices. $D'$ is a matrix representing the corresponding reshaped data.}
\label{data_reshape}
\end{figure*}

Moreover, in order to score the transformed models, we reshape the longitudinal data accordingly. Figure~\ref{data_reshape} shows an illustration of the data reshaping. Suppose we are given longitudinal data with $s$ instances, $p$ variables, and $i$ time slices. We assume that the original data shape is in a form of a matrix $D$ of size $s\times q$, with $q=p\times i$. The reshaped data is then a matrix $D'$ of size $s'\times q'$, with $s'=s(i-1)$ and $q'=2p$. Having such reshaped data allows us to use standard SEM software to compute the scores.

\subsubsection{Constrained SEM}
\label{sec:constrainedSEM}
In practice, we are often given some prior knowledge about the data. The prior knowledge which may be, e.g., results of previous studies, gives us some constraints in terms of causal relations. For example, in the case of, say disease \emph{A}, there exists some \emph{common} knowledge which tells us that symptom \emph{S} does not cause disease \emph{A} directly. In terms of a SEM specification, the prior knowledge can be translated into a constrained SEM in which there is no directed edge from variable $S$ (denotes symptom \emph{S}) to variable $A$ (denotes disease \emph{A}); this still allows for directed edges from $A$ to $S$ or directed paths (indirect relationships) from $S$ to $A$, e.g., a path $S\to...\to A$ with any variables in between. S3C, and hence S3L allow for such prior knowledge to be included in the model. In S3L, this prior knowledge only applies to the intra-slice causal relationships.

Model specifications should comply with any prior knowledge when performing specification search and when measuring the edge and causal path stability. Recall that in order to measure the stability, all optimal models (DAGs) are converted into their corresponding equivalence class models (CPDAGs). This model transformation, however, could result in CPDAGs that are inconsistent with the prior knowledge. For example, a constraint  $A \not\to B$ may be violated since arcs $B \to A$ in the DAG may be converted into undirected (reversible) edges $A-B$ in the CPDAG.
In order to preserve constraints, we therefore extended an efficient DAG-TO-CPDAG algorithm of Chickering\cite{chickering2002learning}, as described in Rahmadi et al\cite{Rahmadi2016}. Essentially, the motivation of our extension to Chickering's algorithm is similar to that of Meek's algorithm\cite{meek1995causal}, that is, to obtain a CPDAG consistent with prior knowledge.

\subsubsection{Estimating causal effects}
\label{causalEffect}
We employ IDA\cite{maathuis2009estimating} to estimate the total causal effects of a covariate $X_i$ on a response $Y$ from the relevant structures. This method works as follows. Given a CPDAG $G=\{G_1,\dotsc,G_m\}$ which contains $m$ different DAGs in its equivalence class, IDA applies intervention calculus \cite{pearl2000causality,pearl2003statistics} to each DAG $G_j$ to obtain multisets $\Theta_i=\{\theta_{ij}\}_{j \in{1,\dotsc,m}}, i=1,\dotsc,p$, where $p$ is the number of covariates. $\theta_{ij}$ specifies the possible causal effect of $X_i$ on $Y$ in graph $G_j$.

Causal effects can be computed using so-called intervention calculus\cite{pearl2000causality}, which aims to determine the amount of change in a response variable $Y$ when one would manipulate the covariate $X_i$ (and not the other variables). Note that this notion differs from a regression-type of association (see IDA paper for illustrative examples). Given a DAG $G_j$, the causal effect $\theta_{ij}$ can be computed using the so-called back-door adjustment, which takes into account the associations between $Y$, $X_i$ and the parents $\mathrm{pa}_i(G_j)$ of $X_i$ in $G_j$. Under the assumption that the distribution of the data is normal and the model is linear, causal effects can be computed from a regression of $Y$ on $X_i$ and its parents. Specifically, we have Maathuis et al.,\cite{maathuis2009estimating} $\theta_{ij} = \beta_{i|\mathrm{pa}_i(G_j)}$,
where, for any set $S \subseteq \{X_1,\dotsc,X_p,Y\}\setminus \{X_i\}$,
\begin{equation*}
\centering
  \beta_{i|S}=
   \begin{cases}
   0, & \mbox{if } Y \in S\\
   \mbox{coefficient of } X_i \mbox{ in } Y\sim X_i+S, & \mbox{if } Y \not\in S,
   \end{cases}
\label{causal_effect}
\end{equation*}
and $Y\sim X_i+S$ is the linear regression of $Y$ on $X_i$ and $S$. Note that IDA estimates the total causal effect from a covariate and response, which considers all possible, either direct or indirect, causal paths from the covariate to the response.

IDA works for continuous, normally distributed variables and then only requires their observed covariance matrix as input to compute the regression coefficients. Following Drasgow \cite{drasgow1988polychoric}, we treat discrete variables as surrogate continuous variables, substituting the polychoric correlation for the correlation between two discrete variables and the polyserial correlation between a discrete and a continuous variable.

Our fitting procedure does not yield a single CPDAG, but a whole set of CPDAGs to represent the given data. We therefore extend IDA as follows. We gather $G_{\pi_{\mathrm{bic}}}$, the CPDAGs of all optimal models with complexity equal to $\pi_{\mathrm{bic}}$. For each CPDAG $G\in G_{\pi_{\mathrm{bic}}}$, we compute the possible causal effects $\Theta$ of each relevant causal path using IDA. For example, for the causal effect from $X$ to $Y$, we obtain estimates $\Theta_{X\to Y}^k, k=1,\dotsc,N$, where $N$ is the number of subsets. All causal effect estimations in $\Theta_{X\to Y}^k$ are then concatenated into a single multiset $\Theta_{X\to Y}$.

To represent the estimated causal effects from $X$ to $Y$, we compute the median $\tilde{\Theta}_{X\to Y}$ and iff $X$ and $Y$ are continuous variables, we standardize the estimation using
\begin{equation*}
\frac{\tilde{\Theta}_{X\to Y}\cdot \sigma_X}{\sigma_Y},
\end{equation*}
where $\sigma_X$ and $\sigma_Y$ are the standard deviations of the covariate and the response, respectively. Standardized causal effects allow us to meaningfully compare them.

\section{Results and discussion}
\label{sec:experimental}
\subsection{Implementation}
We implemented S3C and S3L as an R package named \texttt{stablespec}.
The package is publicly available at the Comprehensive \textsf{R} Archive Network (CRAN)\footnote{\url{https://cran.r-project.org/web/packages/stablespec/index.html}}, so it can be installed directly, e.g., from the \textsf{R} console by typing \texttt{install.package("stablespec")} or from RStudio. We also included a package documentation as a brief tutorial on how to use the functions.

\subsection{Parameter settings}
For application to simulated data and real-world data, we subsampled $50$ and $100$ subsets from the data with size $\lfloor |D|/2 \rfloor$, respectively. We did not do comprehensive parameter tuning for NSGA-{II}, instead, we followed guidelines provided in Grefenstette\cite{grefenstette1986optimization}.
The parameters for applications to both simulated and real-world data were set as follows: the number of iterations was $35$, the number of models in the population was $150$, the probability of applying crossover was $0.85$, the probability of applying mutation to a model structure was $0.07$, and the selection strategy was binary tournament selection\cite{miller1995genetic}.
We score models using the \emph{chi-square} $\chi^2$ and the \emph{model complexity}. The $\chi^2$ is considered the original fit index in SEM and measures how close the model-implied covariance matrix is to the sample covariance matrix\cite{kline2011principles}. The model complexity represents how many parameters (arcs) need to be estimated in the model. The maximum model complexity with $p$ variables is given by $p(p-1)/2$.

When using multi-objective optimization we minimize both the $\chi^2$ and model complexity objectives.
These two objectives are, however, conflicting with each other. For example, minimizing the model complexity typically means compromising the data fit.

\subsection{Application to simulated data}
\subsubsection{Data Generation}
\begin{figure}[!bp]
\centering
\includegraphics[width=0.35\textwidth]{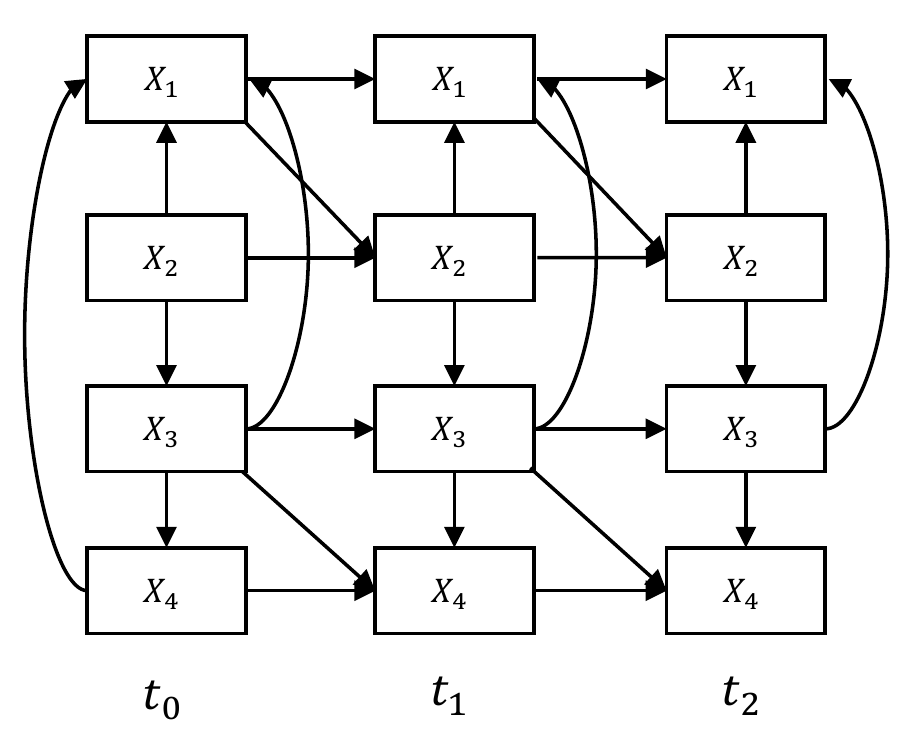}
\caption{The longitudinal model with four variables and three time slices, used to generate simulated data.}
\label{modelAB}
\end{figure}

We generated data sets from a longitudinal model containing four continuous variables and three time slices (depicted by Figure~\ref{modelAB}).
We generated ten data sets for each of these sample sizes: $400$ and $2000$, with random parameterizations. All data sets are made publicly available.\footnote{Available at \url{https://tinyurl.com/smmr-rahmadi-dataset}}

\subsubsection{Performance measure}
We conducted comparisons between S3L with FGES, PC-stable, CPC, CPC-stable, and PC-Max in two different scenarios: with and without prior knowledge about part of the causal directions.
In the case of prior knowledge, we added that variable $X_1$ at $t_i$ cannot cause variables $X_2$ and $X_3$ at $t_i$ directly. This prior knowledge translates to constraints that the various methods can use to restrict their search space. In addition to both scenarios, we also added longitudinal constraints to the models of FGES, PC-stable, CPC, CPC-stable, and PC-Max the same as those used in the transition model of S3L, i.e., there is no intra-causal relationship from time $t_{i-1}$ and the inter-slice causal relationships always go forward in time $t_{i-1}$ to $t_i$.

The parameters of FGES, PC-stable, CPC, CPC-stable, and PC-Max used in this simulation are set following some existing examples\cite{pcalg2012,rcausal2016,maathuis2009estimating}. For FGES, the penalty of BIC score is $2$ and the vertex degree in the forward search is not limited. For PC-stable, CPC, CPC-stable, and PC-Max, the significance level when testing for conditional independence is $0.01$, and the maximum size of the conditioning sets is infinity.

Moreover, as the true model is known, we measure the performance of all approaches by means of the \emph{Receiver Operating Characteristic} (ROC)\cite{fawcett2004roc} for both edges and causal paths.
We compute the \emph{True Positive Rate} (TPR) and the \emph{False Positive Rate} (FPR) based on the CPDAG of the true model. As for example, in the case of edge stability, a true positive means that an edge obtained by our method or the other approaches is present in the CPDAG of the ground truth.

To compare the ROC curves of our method and those of alternative approaches, we employed three significance tests. The first two tests, as introduced in\cite{delong1988comparing} and in\cite{robin2011proc} compare the \emph{Area Under the Curve} (AUC) of the ROC curves by using the theory of U-statistics and bootstrap replicates, respectively. The third test\cite{venkatraman1996distribution} compares the actual ROC curves by evaluating the absolute difference and generating rank-based permutations to compute the statistical significance. The null hypothesis is that (the AUC of) the ROC curves of our method and those of alternative approaches are identical.

Furthermore, we computed the ROC curves using two different schemes: averaging and individual. Both schemes are applied to all methods and to all data sets generated.
In the averaging scheme, the ROC curves are computed from the average edge and causal path stability from different data sets, and then the statistical significance tests are applied to these ROC curves. On the other hand, in the individual scheme the ROC curves are computed from the edge and causal path stability on each data set. We then applied individual statistical significance tests on the ROC curves for each data set and used Fisher's method\cite{fisher1925statistical,10.2307/2681650}, to combine these test results into a single test statistic.

The experimental designs (with and without prior knowledge) and the ROC schemes (averaging and individual) are aimed to show empirically and comprehensively how robust the results are of each approach in various practical cases as well as against changes in the data.

\begin{figure*}[!htbp]
\centering
\mbox{
\subfloat[]
{
\includegraphics[width=0.45\textwidth]{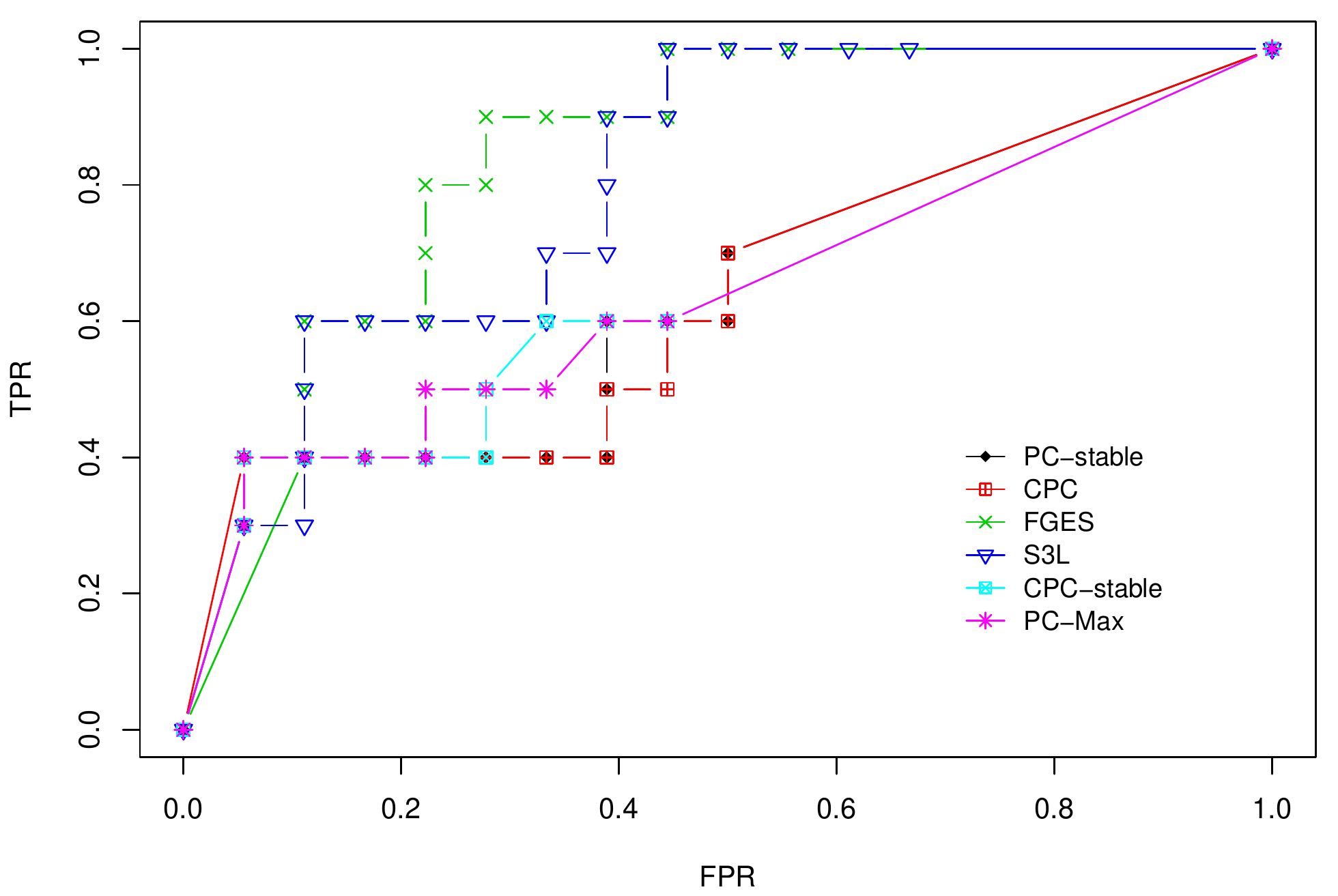}
\label{edgeROC400}
}}
\mbox{%%
\subfloat[]
{
\includegraphics[width=0.45\textwidth]{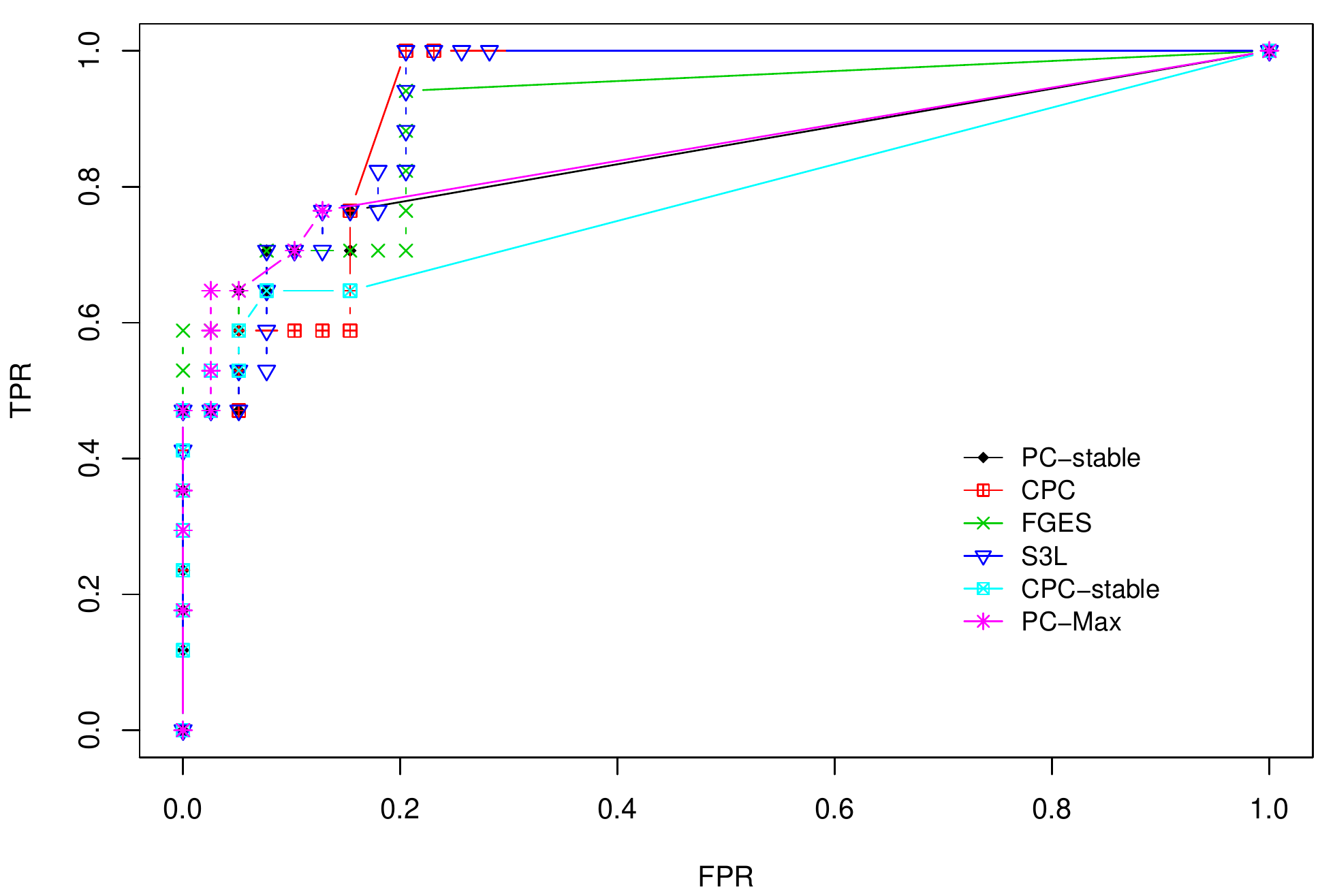}
\label{causalROC400} %causalROC4v3t
}}
\mbox{
\subfloat[]
{
\includegraphics[width=0.45\textwidth]{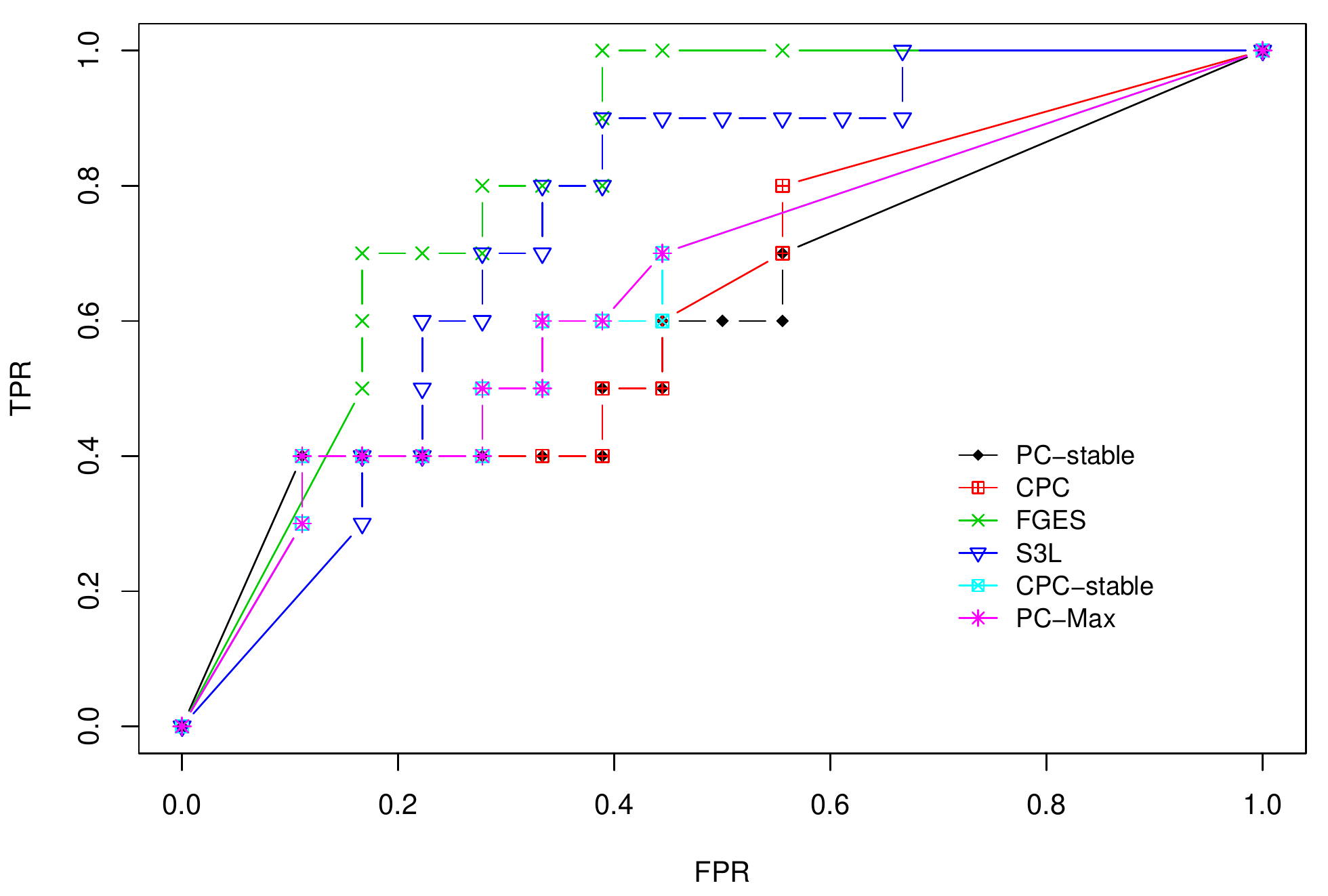}
\label{edgeROC400c}
}}
\mbox{%%
\subfloat[]
{
\includegraphics[width=0.45\textwidth]{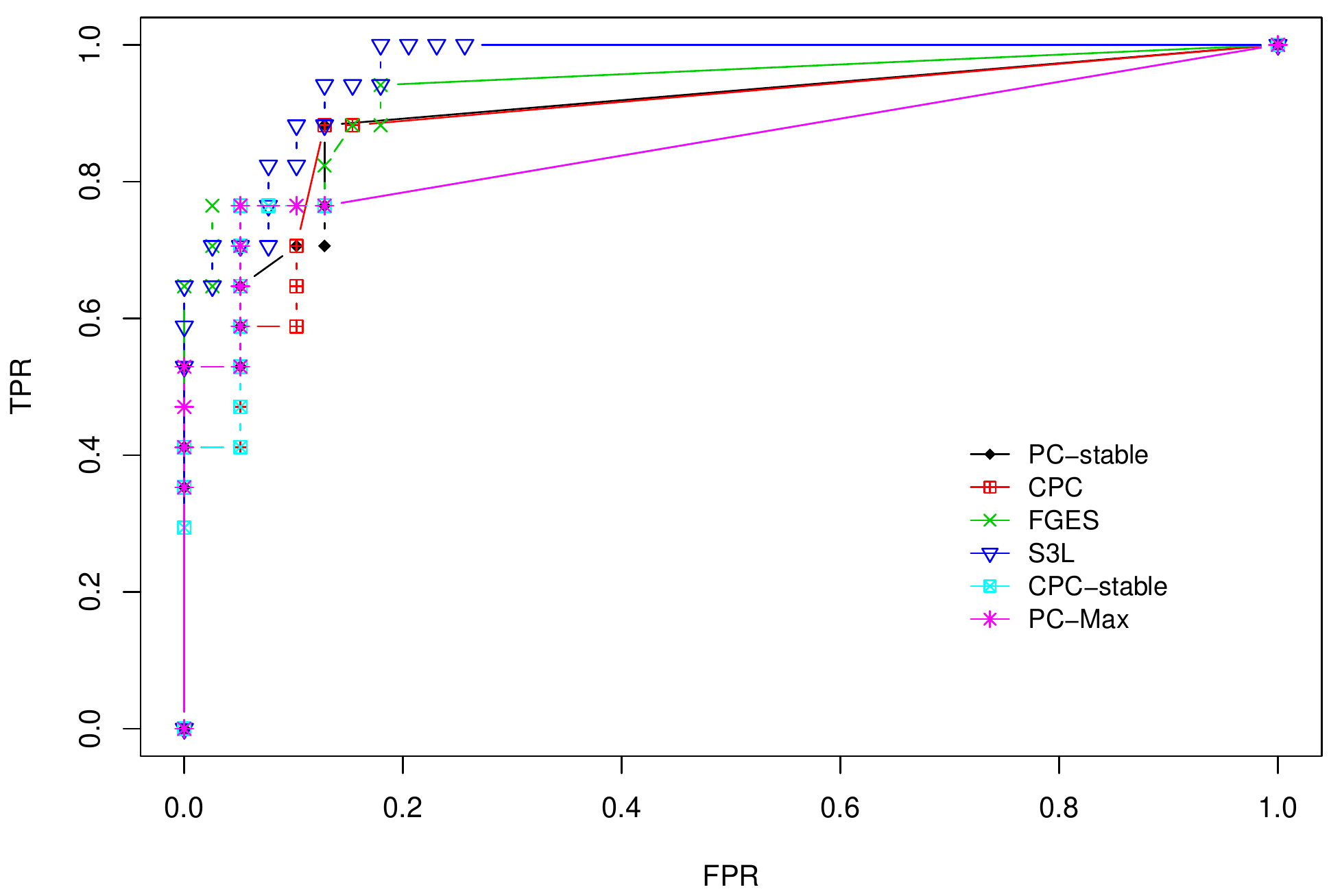}
\label{causalROC400c}
}}
\caption{Results from simulation data with sample size 400: ROC curves for (a) the edge stability and (b) the causal path stability (without prior knowledge), and (c) the edge path stability and (d) the causal path stability (with prior knowledge), for different values of $\pi_{\mathrm{sel}}$ in the range of $[0,1]$. Table~\ref{tableAUC400} lists the corresponding AUCs.}
\label{ROCfor4v}
\end{figure*}
\begin{figure*}[!htbp]
\centering
\mbox{
\subfloat[]
{
\includegraphics[width=0.45\textwidth]{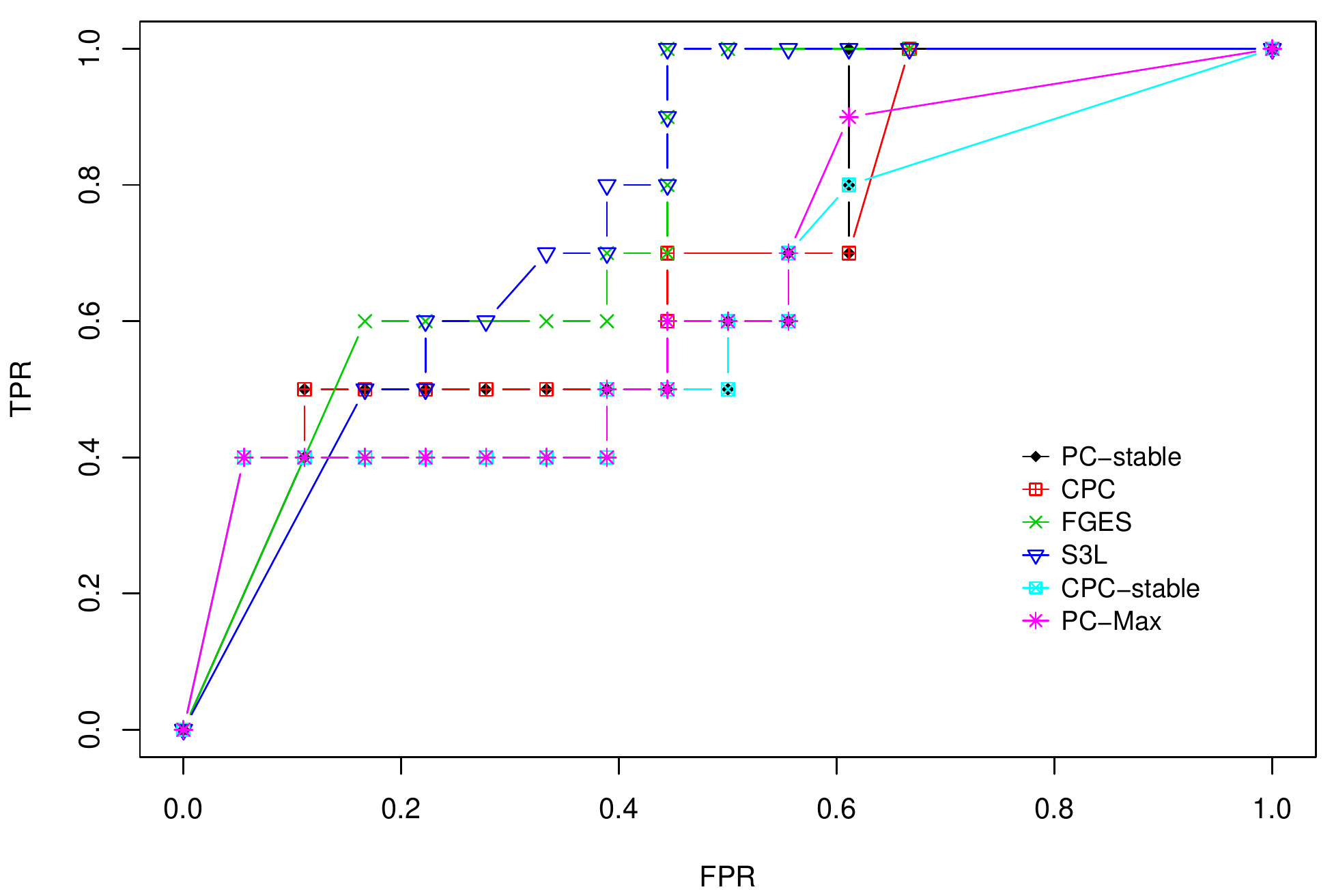}
\label{edgeROC2000}
}}
\mbox{%%
\subfloat[]
{
\includegraphics[width=0.45\textwidth]{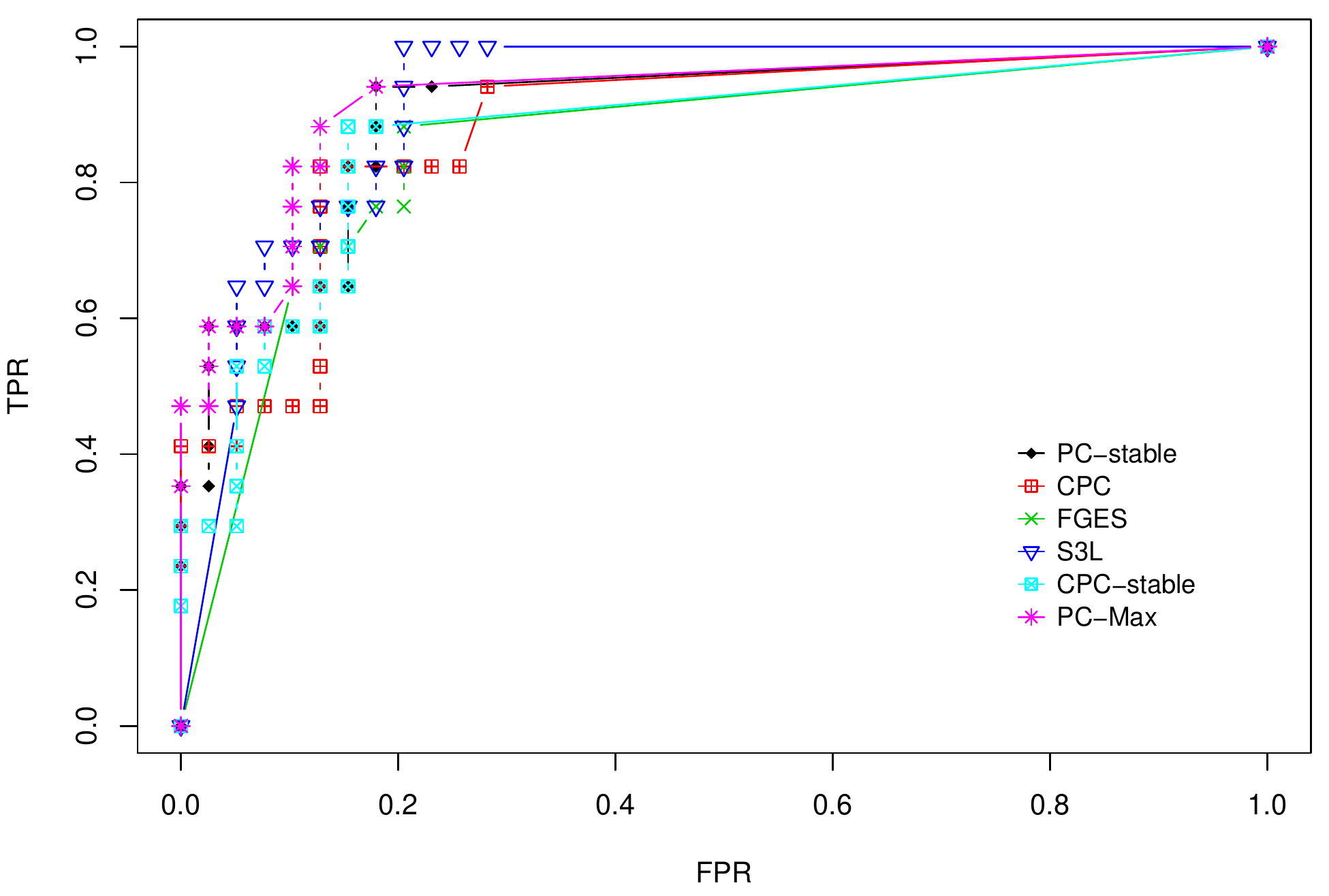}
\label{causalROC2000} %causalROC4v3t
}}
\mbox{
\subfloat[]
{
\includegraphics[width=0.45\textwidth]{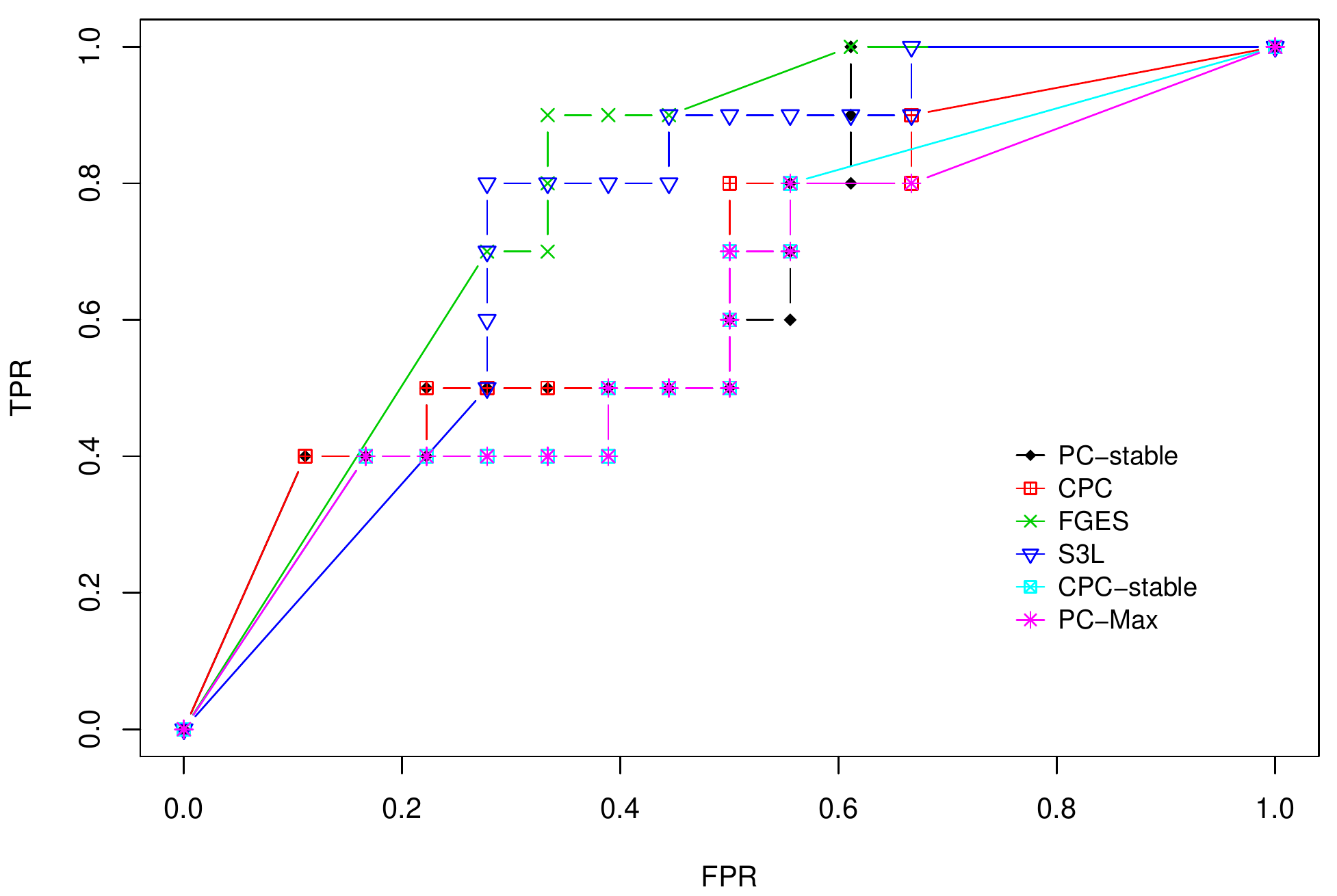}
\label{edgeROC2000c}
}}
\mbox{%%
\subfloat[]
{
\includegraphics[width=0.45\textwidth]{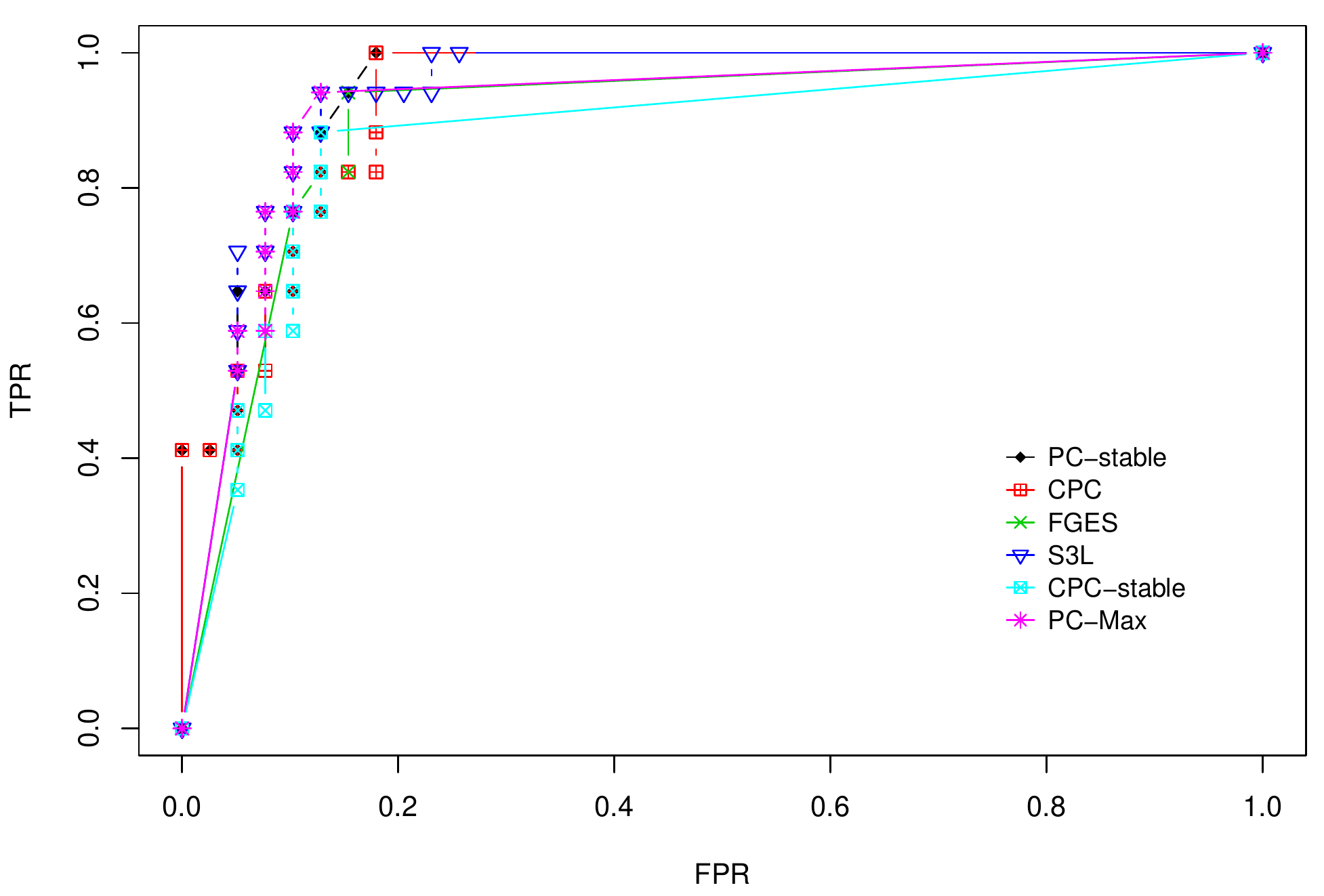}
\label{causalROC2000c}
}}
\caption{Results from simulation data with sample size 2000: ROC curves for (a) the edge stability and (b) the causal path stability (without prior knowledge), and (c) the edge path stability and (d) the causal path stability (with prior knowledge), for different values of $\pi_{\mathrm{sel}}$ in the range of $[0,1]$. Tables~\ref{tableAUC2000} lists the corresponding AUCs.}
\label{ROCfor4v2000}
\end{figure*}
\subsubsection{Discussion}
We first discuss the result of our experiments on the data set with sample size $400$. Figure~\ref{ROCfor4v} shows the ROC curves for the edge stability ((a) and (c)) and the causal path stability ((b) and (d)) from the averaging scheme. Panels (a) and (b) represent the results without prior knowledge, while panels (c) and (d) represent the results with prior knowledge. Table~\ref{tableAUC400} lists the corresponding AUCs.

Tables~\ref{tableROC400} and \ref{tableROC400c} present the results of the significance tests for both the averaging and individual schemes in the experiment with and without prior knowledge, respectively.
In the case without prior knowledge, generally the AUCs of the edge and the causal path stability of S3L are better (\emph{p}-value $\leq 0.05$, or even $\leq 0.001$, few of them are marginally significant, e.g., \emph{p}-value $\leq 0.1$) than those of other approaches according to both schemes, except those of FGES for which generally there is no evidence of a difference (\emph{p}-value $> 0.1$).
In the case with prior knowledge, in general the results are similar to those of experiment without prior knowledge, but now the AUC of the causal path stability of S3L is better (\emph{p}-value $\leq 0.05$) than that of FGES. The ROC of the causal path stability of S3L is now also better (\emph{p}-value $\leq 0.05$) than those of PC-stable, CPC, CPC-stable, and PC-Max according to the individual scheme. This is an improvement over the experiment without prior knowledge.

Next we discuss the result of our experiments on the data set with sample size $2000$. Figure~\ref{ROCfor4v2000} shows the ROC curves and Table~\ref{tableAUC2000} lists the corresponding AUCs.
Tables~\ref{tableROC2000} and \ref{tableROC2000c} list the results of the significance tests for both the averaging and individual schemes in the experiment with and without prior knowledge, respectively. In the case without prior knowledge, generally the AUCs of the edge and the causal path stability of S3L are better than (\emph{p}-value $\leq 0.05$) those of other approaches according to the individual scheme. Moreover, the ROCs of the edge and the causal path stability of S3L are better than those of FGES (\emph{p}-value $\leq 0.001$) and CPC-stable (\emph{p}-value $\leq 0.1$), respectively, according to the individual scheme. In the case with prior knowledge, the results are pretty much similar to those of the experiment without prior knowledge, but only now the \emph{p}-value tends to become smaller, e.g., (\emph{p}-value $\leq 0.001$).

To conclude, we see that in general S3L attains at least comparable performance as, but often a significant improvement over, alternative approaches. This holds in particular for causal directions and in the case of a small sample size. The presence of prior knowledge enhances the performance of the S3L.

\begin{sidewaystable*}%[!htbp]
\centering
\small
\caption{Table of \emph{p}-values from comparisons on data set with sample size $400$ between S3L and alternative approaches without prior knowledge. The null hypothesis is that (the AUC of) the ROC curves of S3L and those of alternative approaches are equivalent. For each significance test, we compared the ROC of the edge (Edge) and causal path (Causal) stability (see Figure~\ref{edgeROC400} and~\ref{causalROC400}) on both averaging (Avg.) and individual (Ind.) schemes.}
\begin{tabular}{|c|c|c|c|c|c|c|c|c|c|c|c|}%{*12c}
\cline{3-12}
\multicolumn{2}{l}{} &
\multicolumn{2}{|c|}{FGES} &
\multicolumn{2}{c|}{PC-stable} &
\multicolumn{2}{c|}{CPC} &
\multicolumn{2}{c|}{CPC-stable} &
\multicolumn{2}{c|}{PC-Max} \\
\hline
\multicolumn{2}{|c|}{Significance test} & Avg. & Ind. & Avg. & Ind. & Avg. & Ind. & Avg. & Ind. & Avg. & Ind.\\
\hline
	{DeLong\cite{delong1988comparing}}
    & {Edge} &
    $0.315$ & $0.909$ &
    $\boldsymbol{0.021}$ & $\boldsymbol{<10^{-5}}$ &
    $\boldsymbol{0.025}$ & $\boldsymbol{<10^{-5}}$ &
    $\textsl{0.052}$ &  $\boldsymbol{<10^{-5}}$ &
    $\boldsymbol{0.050}$ & $\boldsymbol{<10^{-5}}$
    \\
	{} & {Causal} &
    $0.451$ & $0.109$ &
    $\textsl{0.069}$ & $\boldsymbol{<10^{-5}}$ &
    $0.825$ & $\boldsymbol{<10^{-5}}$ &
    $\boldsymbol{0.012}$ & $\boldsymbol{<10^{-5}}$ &
    $0.126$ & $\boldsymbol{<10^{-5}}$
    \\
    \hline
    {Bootstrap\cite{robin2011proc}} &
    {Edge} &
    $0.331$ & $0.935$ &
    $\boldsymbol{0.020}$ & $\boldsymbol{<10^{-5}}$ &
    $\boldsymbol{0.024}$ & $\boldsymbol{<10^{-5}}$ &
    $\textsl{0.051}$ & $\boldsymbol{<10^{-5}}$ &
    $\boldsymbol{0.049}$ & $\boldsymbol{<10^{-5}}$
    \\
    {} & {Causal} &
    $0.466$ & $\textsl{0.090}$ &
    $\textsl{0.063}$ & $\boldsymbol{<10^{-5}}$ &
    $0.830$ & $\boldsymbol{<10^{-5}}$ &
    $\boldsymbol{0.010}$ & $\boldsymbol{<10^{-5}}$ &
    $0.121$ & $\boldsymbol{<10^{-5}}$
    \\
    \hline
	{Venkatraman\cite{venkatraman1996distribution}}
    & {Edge} &
    $0.304$ & $0.906$ &
    $\textsl{0.091}$ & $0.102$ &
    $\textsl{0.076}$ & $0.118$ &
    $0.359$ & $0.516$ &
    $0.449$ & $0.743$
    \\
    {} & {Causal} &
    $0.332$ & $0.197$ &
    $0.831$ & $0.225$ &
    $0.845$ & $0.365$ &
    $0.569$ & $0.512$ &
    $0.584$ & $0.131$
    \\
    \hline
\end{tabular}
\label{tableROC400}

\bigskip\bigskip
\small
\caption{Table of \emph{p}-values from comparisons on data set with sample size $400$ between S3L and alternative approaches with prior knowledge. The null hypothesis is that (the AUC of) the ROC curves of S3L and those of alternative approaches are equivalent. For each significance test, we compared the ROC of the edge (Edge) and causal path (Causal) stability (see Figure~\ref{edgeROC400c} and~\ref{causalROC400c}) on both averaging (Avg.) and individual (Ind.) schemes.}
\begin{tabular}{|c|c|c|c|c|c|c|c|c|c|c|c|}%{*12c}
\cline{3-12}
\multicolumn{2}{l}{} &  \multicolumn{2}{|c|}{FGES} & \multicolumn{2}{c|}{PC-stable} & \multicolumn{2}{c|}{CPC} & \multicolumn{2}{c|}{CPC-stable} & \multicolumn{2}{c|}{PC-Max}\\
\hline
\multicolumn{2}{|c|}{Significance test} & Avg. & Ind. & Avg. & Ind. & Avg. & Ind. & Avg. & Ind. & Avg. & Ind.\\
\hline
	{DeLong\cite{delong1988comparing}} & {Edge} &
    $\textsl{0.090}$ & $0.146$&
    $\textsl{0.086}$ & $\boldsymbol{<10^{-3}}$ &
    $\textsl{0.099}$ & $\boldsymbol{<10^{-5}}$ &
    $0.219$ & $\boldsymbol{0.001}$ &
    $0.227$ & $\boldsymbol{0.002}$\\
								{} & {Causal} &
    $0.264$ & $\boldsymbol{0.003}$ &
    $\textsl{0.061}$ & $\boldsymbol{<10^{-5}}$ &
    $\boldsymbol{0.035}$ & $\boldsymbol{<10^{-5}}$ &
    $\boldsymbol{0.022}$ & $\boldsymbol{<10^{-5}}$ &
    $\boldsymbol{0.031}$ & $\boldsymbol{<10^{-5}}$\\ \hline

	{Bootstrap\cite{robin2011proc}} & {Edge} &
    $0.118$ & $0.188$ &
    $\textsl{0.084}$ & $\boldsymbol{<10^{-5}}$ &
    $\textsl{0.099}$ & $\boldsymbol{<10^{-5}}$ &
    $0.208$ & $\boldsymbol{<10^{-3}}$ &
    $0.223$ & $\boldsymbol{0.001}$\\
                                {} & {Causal} &
    $0.251$ & $\boldsymbol{0.002}$ &
    $\textsl{0.060}$ & $\boldsymbol{<10^{-5}}$ &
    $\boldsymbol{0.031}$ & $\boldsymbol{<10^{-5}}$ &
    $\boldsymbol{0.020}$ & $\boldsymbol{<10^{-5}}$ &
    $\boldsymbol{0.026}$ & $\boldsymbol{<10^{-5}}$ \\ \hline

	{Venkatraman\cite{venkatraman1996distribution}}
    & {Edge} &
    $0.430$ & $0.598$ &
    $\textsl{0.056}$ & $0.680$ &
    $0.103$ & $0.543$ &
    $0.680$ & $0.998$ &
    $0.707$ & $0.998$\\
    {} & {Causal} &
    $0.637$ & $0.783$ &
    $0.485$ & $\boldsymbol{0.004}$ &
    $\textsl{0.069}$ & $\boldsymbol{<10^{-3}}$ &
    $0.116$ & $\textsl{0.094}$ &
    $0.171$ & $\boldsymbol{0.007}$ \\
                               \hline
\end{tabular}
\label{tableROC400c}

\bigskip\bigskip
\small
\caption{Table of AUCs for the edge and causal path stability for each method, from simulation on data with sample size $400$, with (yes) and without prior knowledge (no).}
\begin{tabular}{|c|c|c|c|c|c|c|c|c|c|c|c|c|}
 \cline{2-13}
   \multicolumn{1}{l}{}& \multicolumn{2}{|c|}{S3L} & \multicolumn{2}{|c}{FGES} & \multicolumn{2}{|c}{PC-stable} & \multicolumn{2}{|c}{CPC} & \multicolumn{2}{|c}{CPC-stable} & \multicolumn{2}{|c|}{PC-Max} \\ \hline
  AUC & no & yes & no & yes & no & yes & no & yes & no & yes & no & yes \\ \hline
  edge & $0.80$ & $0.74$ & $0.83$ & $0.81$ & $0.63$ & $0.60$ & $0.63$ & $0.62$ & $0.63$ & $0.65$ & $0.63$ & $0.65$ \\ \hline
  causal & $0.92$ & $0.96$ & $0.90$ & $0.93$ & $0.84$ & $0.90$ & $0.92$ & $0.89$ & $0.78$ & $0.84$ & $0.85$ & $0.85$ \\ \hline
\end{tabular}
\label{tableAUC400}
\end{sidewaystable*}

\begin{sidewaystable*}%[!htbp]
\centering
\small
\caption{Table of \emph{p}-values from comparisons on data set with sample size $2000$ between S3L and alternative approaches without prior knowledge. The null hypothesis is that (the AUC of) the ROC curves of S3L and those of alternative approaches are equivalent. For each significance test, we compared the ROC of the edge (Edge) and causal path (Causal) stability (see Figure~\ref{edgeROC2000} and~\ref{causalROC2000}) on both averaging (Avg.) and individual (Ind.) schemes.}
\begin{tabular}{|c|c|c|c|c|c|c|c|c|c|c|c|}%{*12c}
\cline{3-12}
\multicolumn{2}{l}{} &  \multicolumn{2}{|c|}{FGES} & \multicolumn{2}{c|}{PC-stable} & \multicolumn{2}{c|}{CPC} & \multicolumn{2}{c|}{CPC-stable} & \multicolumn{2}{c|}{PC-Max}\\
\hline
\multicolumn{2}{|c|}{Significance test} & Avg. & Ind. & Avg. & Ind. & Avg. & Ind. & Avg. & Ind. & Avg. & Ind.\\
\hline
	{DeLong\cite{delong1988comparing}} & {Edge} &
        $1.000$ & $\textsl{0.099}$ &
        $0.223$ & $\boldsymbol{0.010}$ &
        $0.320$ & $\boldsymbol{0.014}$ &
        $\textsl{0.071}$ & $\boldsymbol{0.001}$ &
        $0.118$ & $\boldsymbol{0.009}$ \\
		{} & {Causal} &
        $\textsl{0.052}$ & $\boldsymbol{<10^{-5}}$ &
        $0.563$ & $\boldsymbol{<10^{-3}}$ &
        $0.353$ & $\boldsymbol{<10^{-5}}$ &
        $0.221$ & $\boldsymbol{<10^{-5}}$ &
        $0.952$ & $0.183$\\ \hline

	{Bootstrap\cite{robin2011proc}} & {Edge} &
        $1.000$ & $0.103$ &
        $0.222$ & $\boldsymbol{0.007}$ &
        $0.321$ & $\boldsymbol{0.010}$ &
        $\textsl{0.077}$ & $\boldsymbol{0.003}$ &
        $0.103$ & $\boldsymbol{0.006}$\\

        {} & {Causal} &
         $\boldsymbol{0.045}$ & $\boldsymbol{<10^{-5}}$ &
         $0.554$ & $\boldsymbol{<10^{-3}}$ &
         $0.357$ & $\boldsymbol{<10^{-5}}$ &
         $0.202$ & $\boldsymbol{<10^{-5}}$ &
         $0.952$ & $0.161$ \\ \hline

	{Venkatraman\cite{venkatraman1996distribution}} &
        {Edge} &
        $0.480$ & $0.963$ &
        $0.187$ & $0.801$ &
        $0.212$ & $0.872$ &
        $\textsl{0.069}$ & $0.900$ &
        $0.100$ & $0.972$\\
     {} & {Causal} &
        $0.418$ & $\boldsymbol{<10^{-3}}$ &
        $0.404$ & $0.637$ &
        $0.289$ & $0.339$ &
        $0.726$ & $0.897$&
        $0.520$ & $0.250$ \\ \hline
\end{tabular}
\label{tableROC2000}

\bigskip\bigskip
\small
\caption{Table of \emph{p}-values from comparisons on data set with sample size $2000$ between S3L and alternative approaches with prior knowledge. The null hypothesis is that (the AUC of) the ROC curves of S3L and those of alternative approaches are equivalent. For each significance test, we compared the ROC of the edge (Edge) and causal path (Causal) stability (see Figure~\ref{edgeROC2000c} and~\ref{causalROC2000c}) on both averaging (Avg.) and individual (Ind.) schemes.}
\begin{tabular}{|c|c|c|c|c|c|c|c|c|c|c|c|}%{*12c}
\cline{3-12}
\multicolumn{2}{l}{} &  \multicolumn{2}{|c|}{FGES} & \multicolumn{2}{c|}{PC-stable} & \multicolumn{2}{c|}{CPC} & \multicolumn{2}{c|}{CPC-stable} & \multicolumn{2}{c|}{PC-Max}\\
\hline
\multicolumn{2}{|c|}{Significance test} & Avg. & Ind. & Avg. & Ind. & Avg. & Ind. & Avg. & Ind. & Avg. & Ind.\\
\hline
	{DeLong\cite{delong1988comparing}} & {Edge} &
        $0.296$ & $0.978$
        &$0.413$ & $\boldsymbol{<10^{-3}}$ &
        $0.348$ & $\boldsymbol{0.005}$ &
        $0.147$ & $\boldsymbol{<10^{-3}}$ &
        $0.122$ & $\boldsymbol{<10^{-3}}$ \\
								
        {} & {Causal} &
        $0.142$ & $\boldsymbol{<10^{-5}}$ &
        $0.817$ & $\boldsymbol{<10^{-3}}$ &
        $0.698$ & $\boldsymbol{<10^{-3}}$ &
        $\boldsymbol{0.043}$ & $\boldsymbol{<10^{-5}}$ &
        $0.279$ & $\boldsymbol{<10^{-5}}$\\ \hline

	{Bootstrap\cite{robin2011proc}} &

        {Edge} &
        $0.295$ & $0.983$ &
        $0.412$ & $\boldsymbol{<10^{-3}}$ &
        $0.344$ & $\boldsymbol{0.002}$ &
        $0.146$ & $\boldsymbol{<10^{-5}}$ &
        $0.125$ & $\boldsymbol{<10^{-5}}$\\

        {} & {Causal} &
        $0.144$ & $\boldsymbol{<10^{-5}}$ &
        $0.833$ & $\boldsymbol{<10^{-3}}$ &
        $0.706$ & $\boldsymbol{<10^{-3}}$ &
        $\boldsymbol{0.043}$ & $\boldsymbol{<10^{-5}}$ &
        $0.279$ & $\boldsymbol{<10^{-5}}$ \\ \hline

	{Venkatraman\cite{venkatraman1996distribution}} & {Edge} &
        $0.761$ & $0.862$ &
        $0.119$ & $0.207$ &
        $0.210$ & $0.290$ &
        $0.146$ & $\textsl{0.082}$ &
        $0.140$ & $0.257$\\

        {} & {Causal} &
         $0.486$ & $0.595$ &
         $0.384$ & $0.763$ &
         $0.172$ & $0.742$ &
         $0.488$ & $0.984$ &
         $0.652$ & $0.903$ \\ \hline
\end{tabular}
\label{tableROC2000c}

\bigskip\bigskip
\small
\caption{Table of AUCs for the edge and causal path stability for each method, from simulation on data with sample size $2000$, with (yes) and without prior knowledge (no).}
\begin{tabular}{|c|c|c|c|c|c|c|c|c|c|c|c|c|}
 \cline{2-13}
   \multicolumn{1}{l}{}& \multicolumn{2}{|c|}{S3L} & \multicolumn{2}{|c}{FGES} & \multicolumn{2}{|c}{PC-stable} & \multicolumn{2}{|c}{CPC} & \multicolumn{2}{|c}{CPC-stable} & \multicolumn{2}{|c|}{PC-Max} \\ \hline
  AUC & no & yes & no & yes & no & yes & no & yes & no & yes & no & yes \\ \hline
  edge &
  $0.77$ & $0.73$ &
  $0.77$ & $0.78$ &
  $0.67$ & $0.67$ &
  $0.69$ & $0.65$ &
  $0.62$ & $0.61$ &
  $0.65$ & $0.60$ \\ \hline
  causal &
  $0.92$ & $0.93$ &
  $0.85$ & $0.90$ &
  $0.90$ & $0.94$ &
  $0.88$ & $0.93$ &
  $0.87$ & $0.87$ &
  $0.92$ & $0.91$
    \\ \hline
\end{tabular}
\label{tableAUC2000}
\end{sidewaystable*}

\subsection{Application to real-world data}
Here the true model is unknown, so we can only compare the results of S3L with those reported in earlier studies and interpretation by medical experts. We set the thresholds to $\pi_{\mathrm{sel}} = 0.6$ and $\pi_{\mathrm{bic}}$ to the model complexity where the minimum average of BIC scores is found. By thresholding we get the relevant causal relationships: those which occur in the relevant region. Details of the procedure are given in Section~\ref{sec:stableSpecSearch}.

The model assumptions in the application to real-world data follow from the assumptions of S3L in the default setting. The assumptions include iid samples on each time slice, linear system, independent gaussian noise, no latent variables, stationary, and fairly uniform time intervals between time slices.

Moreover, there is an important note related to the visualization of the stability graphs. A DAG without edges will always be transformed into a CPDAG without edges. A fully connected DAG without prior knowledge will be transformed into a CPDAG with only undirected edges. However if prior knowledge is added, a fully connected DAG will be transformed into a CPDAG in which the edges corresponding to the prior knowledge are directed. From these observations it follows that in the edge stability graph all paths start with a selection probability of $0$ and end up in a selection probability of $1$. In the causal path stability graph when no prior knowledge has been added all paths start with a selection probability of 0 and end up in a selection probability of $0$. However, when prior knowledge is added some of the paths may end up in a selection probability of $1$ because of the added constraints.

\subsubsection{Application to chronic fatigue syndrome data}
Our first application to real-world data considers a longitudinal data set of $183$ patients with \emph{chronic fatigue syndrome} (CFS) who received \emph{cognitive behavior therapy} (CBT)\cite{heins2013process}. Empirical studies have shown that CBT can significantly reduce fatigue severity. In this study we focus on the causal relationships between cognitions and behavior in the process of reducing subject's fatigue severity. We therefore include six variables namely \emph{fatigue} severity, the sense of \emph{control} over fatigue, \emph{focusing} on the symptoms, the objective activity of the patient (\emph{oActivity}), the subject's perceived activity (\emph{pActivity}), and the physical \emph{functioning}. The data set consists of five time slices where
the first and the fifth time slices are the pre- and post-treatment observations, respectively, and the second until the fourth time slices are observations during the treatment. The missing data is $8.7\%$ and to impute the missing values, we used single imputation with \emph{Expectation Maximization} (EM) in SPSS\cite{spss}. As all of the variables have large scales, e.g., in the range between $0$ to $155$, we treat them as continuous variables.
We added prior knowledge that the variable \emph{fatigue} at $t_0$ and $t_i$ does not cause any of the other variables directly. This is a common assumption made in the analysis of CBT in order to investigate the causal impact on fatigue severity\cite{vercoulen1998persistence,heins2013process}.

\begin{figure*}
\centering
\mbox{\subfloat[]
{
\includegraphics[width=0.48\textwidth]{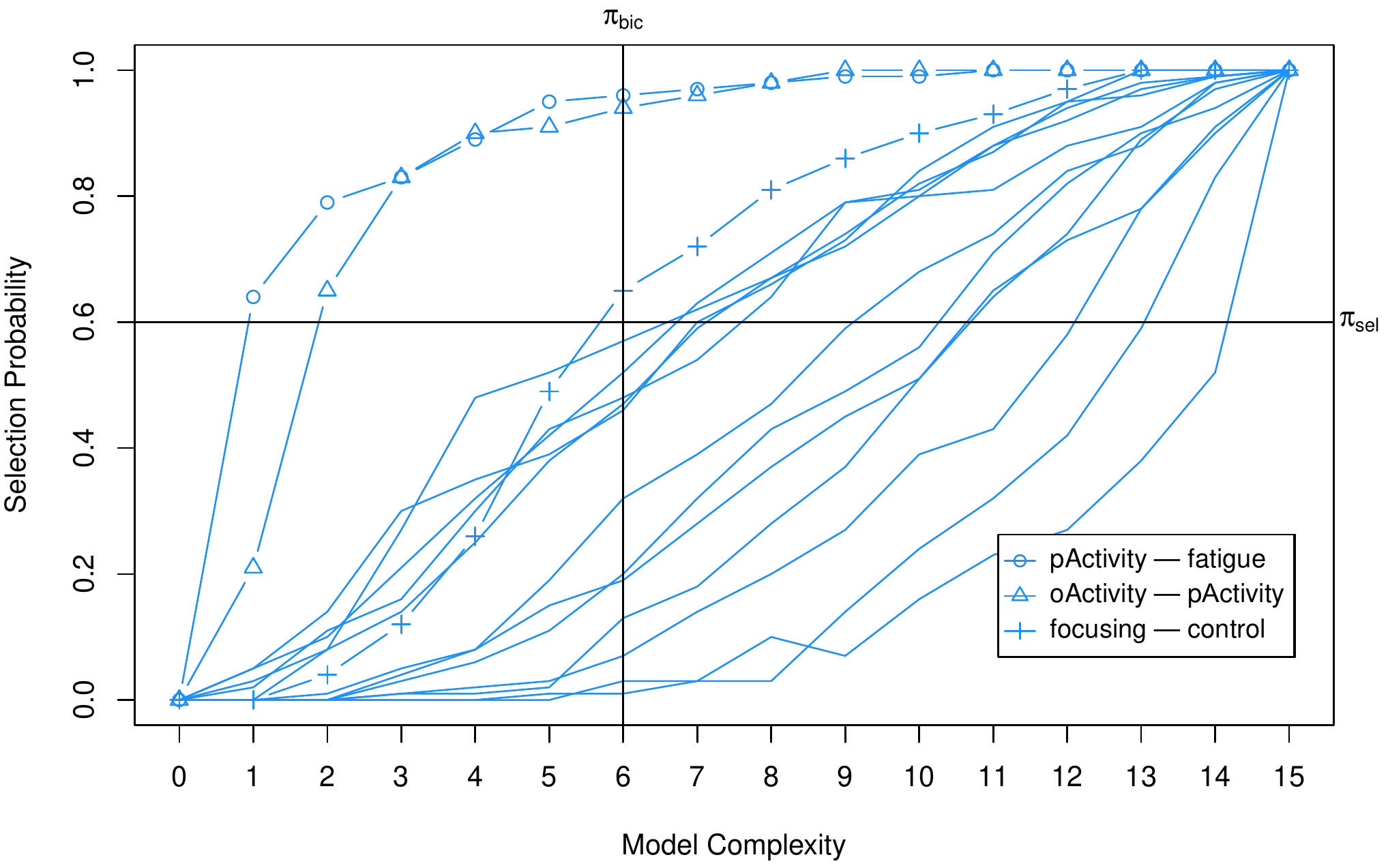}
\label{edgeStabCFSPrior}
}}
\mbox{\subfloat[]
{
\includegraphics[width=0.48\textwidth]{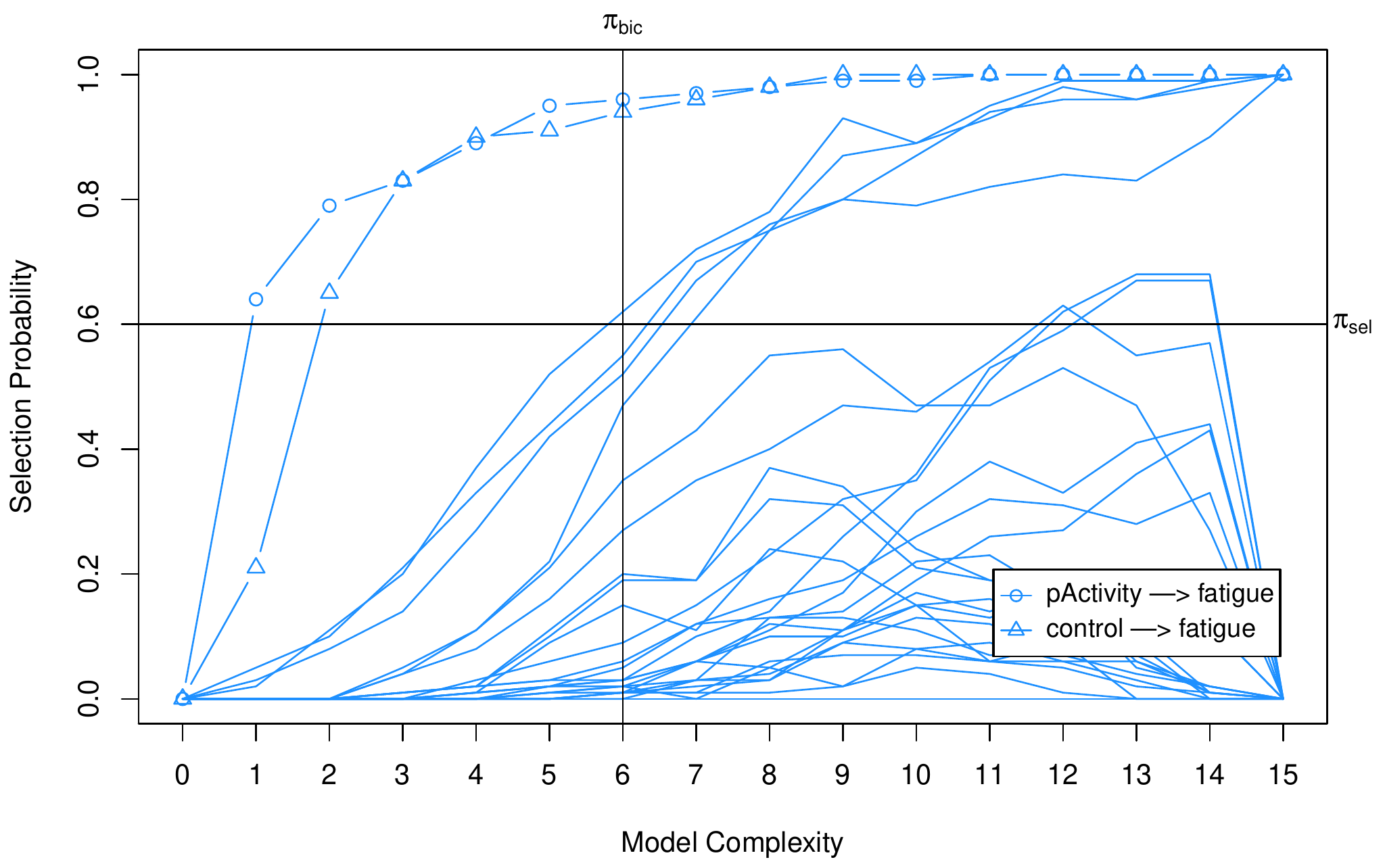}
\label{causalStabCFSPrior}
}}
\mbox{\subfloat[]
{
\includegraphics[width=0.48\textwidth]{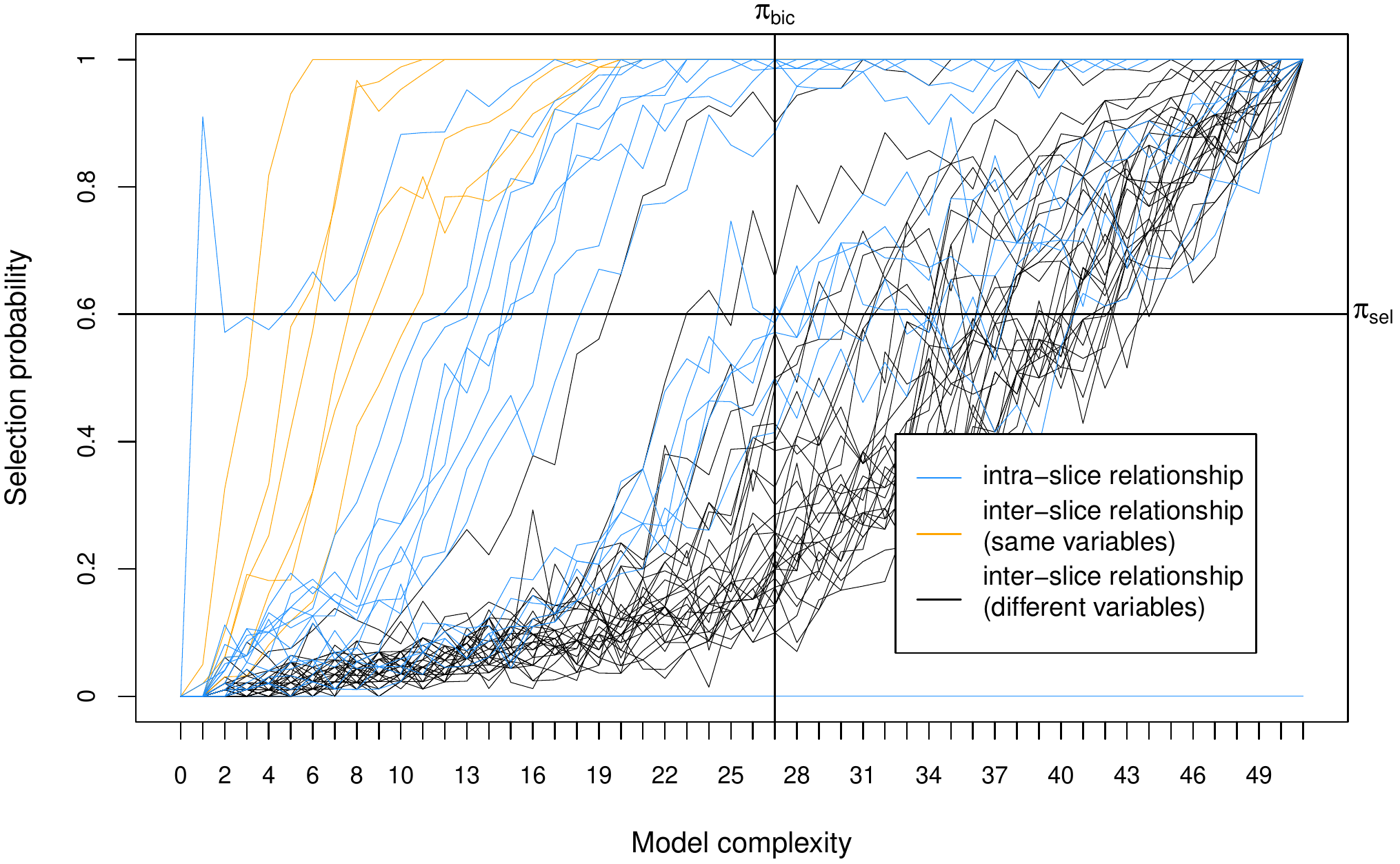}
\label{edgeStabCFSTransition}
}}
\mbox{\subfloat[]
{
\includegraphics[width=0.48\textwidth]{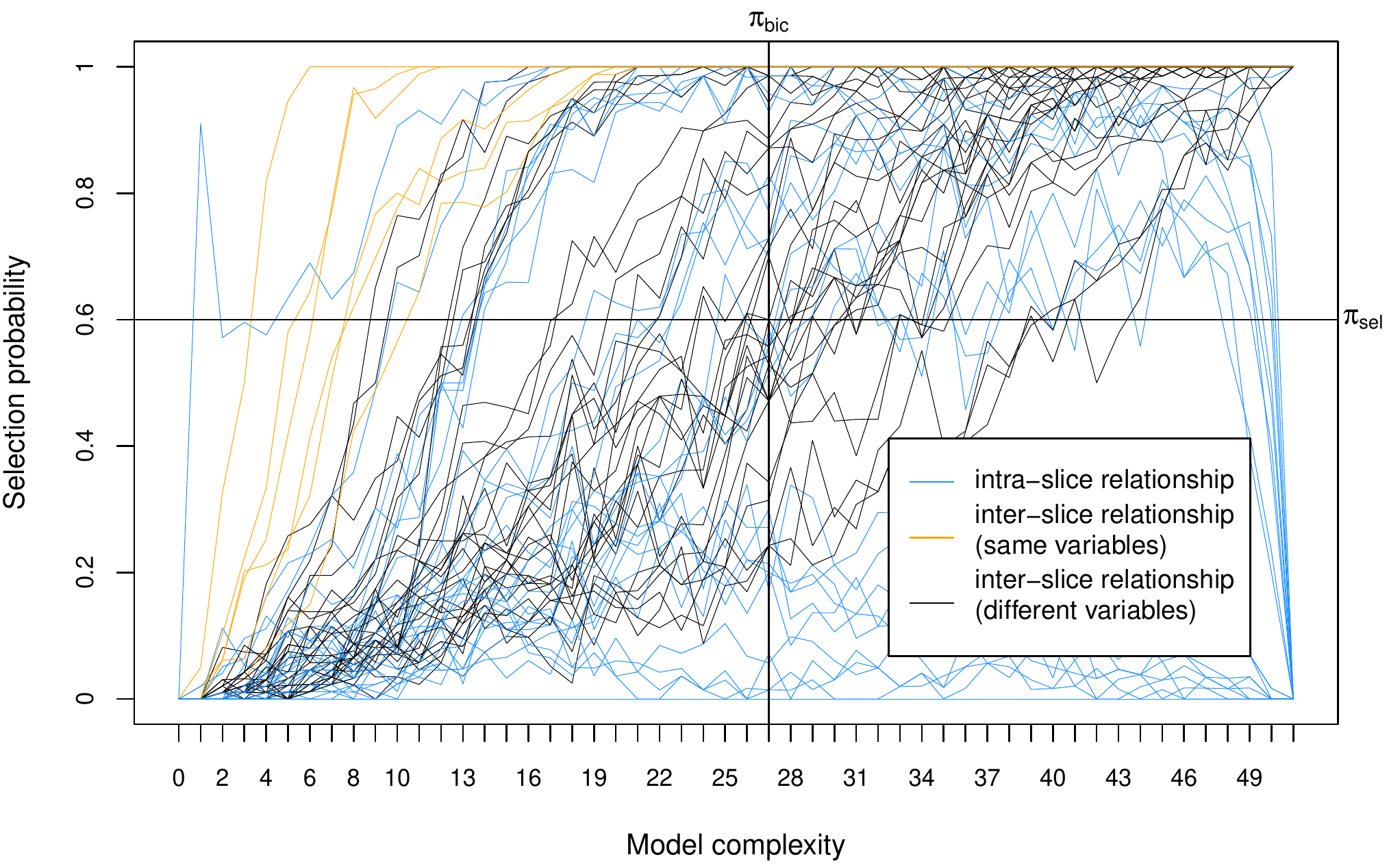}
\label{causalStabCFSTransition}
}}
\caption{The stability graphs of the baseline model in (a) and (b) and the transition model in (c) and (d) for chronic fatigue syndrome, with edge stability in (a) and (c), and causal path stability in (b) and (d). The relevant regions, above $\pi_{\mathrm{sel}}$ and left of $\pi_{\mathrm{bic}}$, contain the relevant structures.
}
\label{stabGraphCFS}
\end{figure*}

\begin{figure*}
\centering
\mbox{\subfloat[]
{
\includegraphics[width=0.19\textwidth]{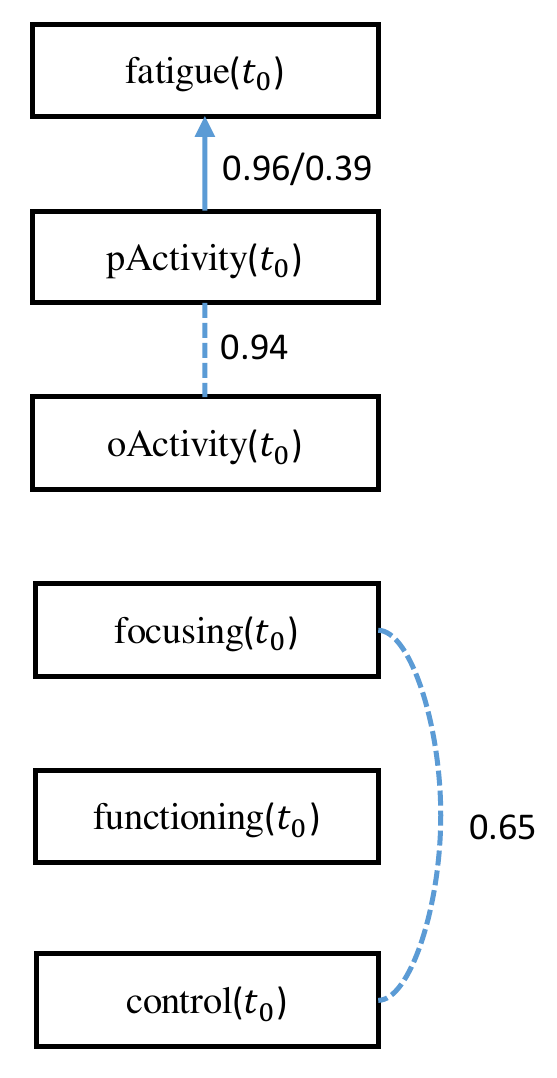}
\label{priorModelCFS}
}}
\qquad
\mbox{\subfloat[]
{
\includegraphics[width=0.675\textwidth]{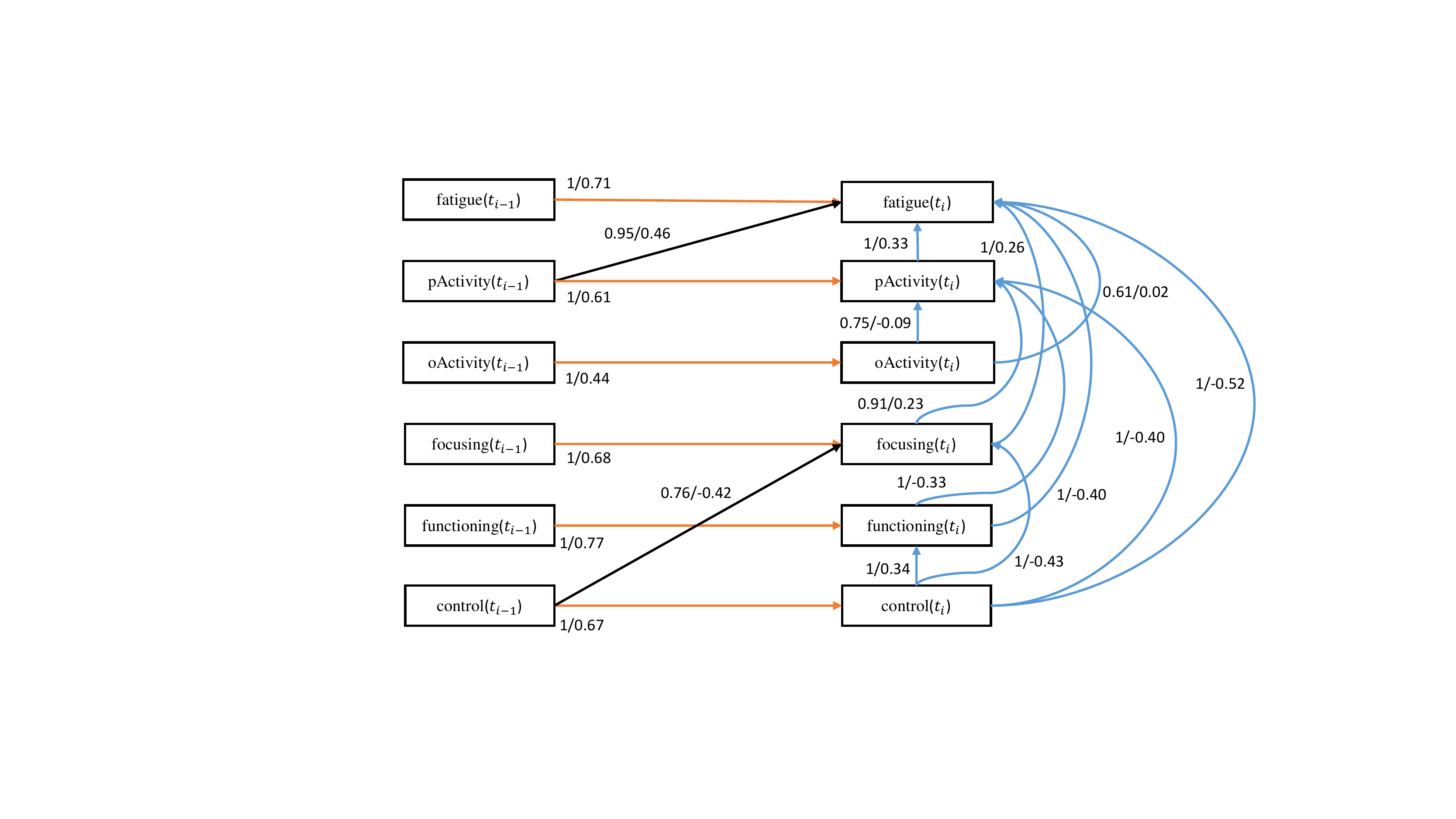}
\label{transitionModelCFS}
}}
\caption{(a) The baseline model and (b) the transition model of chronic fatigue syndrome. The dashed line represents a strong relation between two variables but the causal direction cannot be determined from the data.
Each edge has a reliability score (the highest selection probability in the relevant region of the edge stability graph) and a standardized total causal effect estimation. For example, the annotation \quotes{$1/0.71$} represents a reliability score of $1$ and a standardized total causal effect of $0.71$. Note that the standardized total causal effect represents not just the direct causal effect corresponding to the edge, but the total causal effect also including indirect effects}.
\label{causalModelCFS}
\end{figure*}

First we discuss the baseline model, which only considers the baseline causal relationships.
The corresponding stability graphs can be seen in Figures~\ref{edgeStabCFSPrior} and ~\ref{causalStabCFSPrior}. As mentioned before, $\pi_{\mathrm{sel}}$ is set to $0.6$ and from the search phase of S3L we found that $\pi_{\mathrm{bic}}=6$. Figures~\ref{edgeStabCFSPrior} and ~\ref{causalStabCFSPrior} show that three relevant edges and two relevant causal paths were found.
Following the visualization procedure (see visualization phase in Section~\ref{sec:stableSpecSearch}), we get a baseline model in Figure~\ref{priorModelCFS}.
The model shows that \emph{pActivity} is a direct cause for \emph{fatigue} severity. This follows from the prior assumption that we made and is consistent with earlier works\cite{vercoulen1998persistence,heins2013process}. This causal relationship suggests that a reduction of (perceived) activity, leads to an increase of fatigue. In addition we found a strong relationship between \emph{pActivity} and \emph{oActivity} whose direction cannot be determined. This relationship is somewhat sensible as both variables measuring patient's activity. We also found a connection between \emph{focusing} and \emph{control}, which is not surprising as focusing on symptoms also depends on patient's sense of control over fatigue. One would expect that if a patient has less control on the fatigue, the focus on the symptom would increase.

Next we discuss the transition model, which considers all causal relationships over time slices. The corresponding stability graphs are depicted in Figures~\ref{edgeStabCFSTransition} and \ref{causalStabCFSTransition}. We set $\pi_{\mathrm{sel}}=0.6$ and the search phase of S3L yielded $\pi_{\mathrm{bic}}=27$.
Figures~\ref{edgeStabCFSTransition} shows that nineteen relevant edges were found, consisting of eleven intra-slice (blue lines) and eight inter-slice relationships of which six are \mbox{between} the same variables (orange lines) and two are between different variables (black lines). Figure \ref{causalStabCFSTransition} shows that thirty-five relevant causal paths were found, consisting of twelve intra-slice (blue lines) and twenty-three inter-slice relationships of which six are \mbox{between} the same variables (orange lines) and seventeen are between different variables (black lines).
Applying the visualization procedure, we get the transition model in Figure~\ref{transitionModelCFS}.
The model shows that all variables have intra-slice causal relationships to \emph{fatigue} severity. These relationships are consistent with \cite{vercoulen1998persistence,heins2013process,wiborg2012towards} which conclude that during the CBT an increase in sense of control over fatigue, physical functioning, and perceived physical activity, together with a decrease in focusing on symptoms lead to a lower level of fatigue severity. Interestingly, the actual activity seems insufficient to reduce fatigue severity\cite{heins2013process}, however, how the patient perceives his own activity does seem to help. Additionally, we also found that, with similar causal effects, all variables (except \emph{pActivity} and \emph{fatigue}) also cause the change in fatigue indirectly via \emph{pActivity} as an intermediate variable. This suggests that, as discussed in\cite{heins2013process} an increase in perceived activity does seem important to explain the change in fatigue. The variables \emph{focusing} and \emph{functioning} also appear to be indirect causes of changes in the level of fatigue severity.

\subsubsection{Application to Alzheimer's disease data}
For the second application to real-world data, we consider a longitudinal data set about \emph{Alzheimer's Disease} (AD), which is provided by the \emph{Alzheimer's Disease Neuroimaging Initiative} (ADNI)\cite{weiner2010alzheimer}, and can be accessed at \url{adni.loni.usc.edu}.
The ADNI was launched in 2003 as a public-private partnership, led by Principal Investigator Michael W. Weiner, MD. The primary goal of ADNI has been to test whether serial magnetic resonance imaging (MRI), positron emission tomography (PET), other biological markers, and clinical and neuropsychological assessment can be combined to measure the progression of \emph{mild cognitive impairment} (MCI) and early AD. For up-to-date information see \url{www.adni-info.org}.

In the present paper we focus on patients with MCI, an intermediate clinical stage in AD\cite{petersen1999mild}. Following Haight et al.,\cite{haight2012relative} we include only the variables: subject's cognitive dysfunction (\emph{ADAS-Cog}), hippocampal volume (\emph{hippocampal\_vol}), whole brain volume (\emph{brain\_vol}), and brain glucose metabolism (\emph{brain\_glucose}). The data set contains $179$ subjects with four continuous variables and six time slices. The first time slice captures baseline observations and the next time slices are for the follow-up observations. The missing data is $22.9\%$ and like in the application to CFS, we imputed the missing values using single imputation with EM. We added prior knowledge that the variable \emph{ADAS-Cog} at $t_0$ and $t_i$ does not cause any of the other variables directly.
We performed the search over $100$ subsamples of the original data set.
\begin{figure*}[!htbp]
\centering
\mbox{\subfloat[]
{
\includegraphics[width=0.48\textwidth]{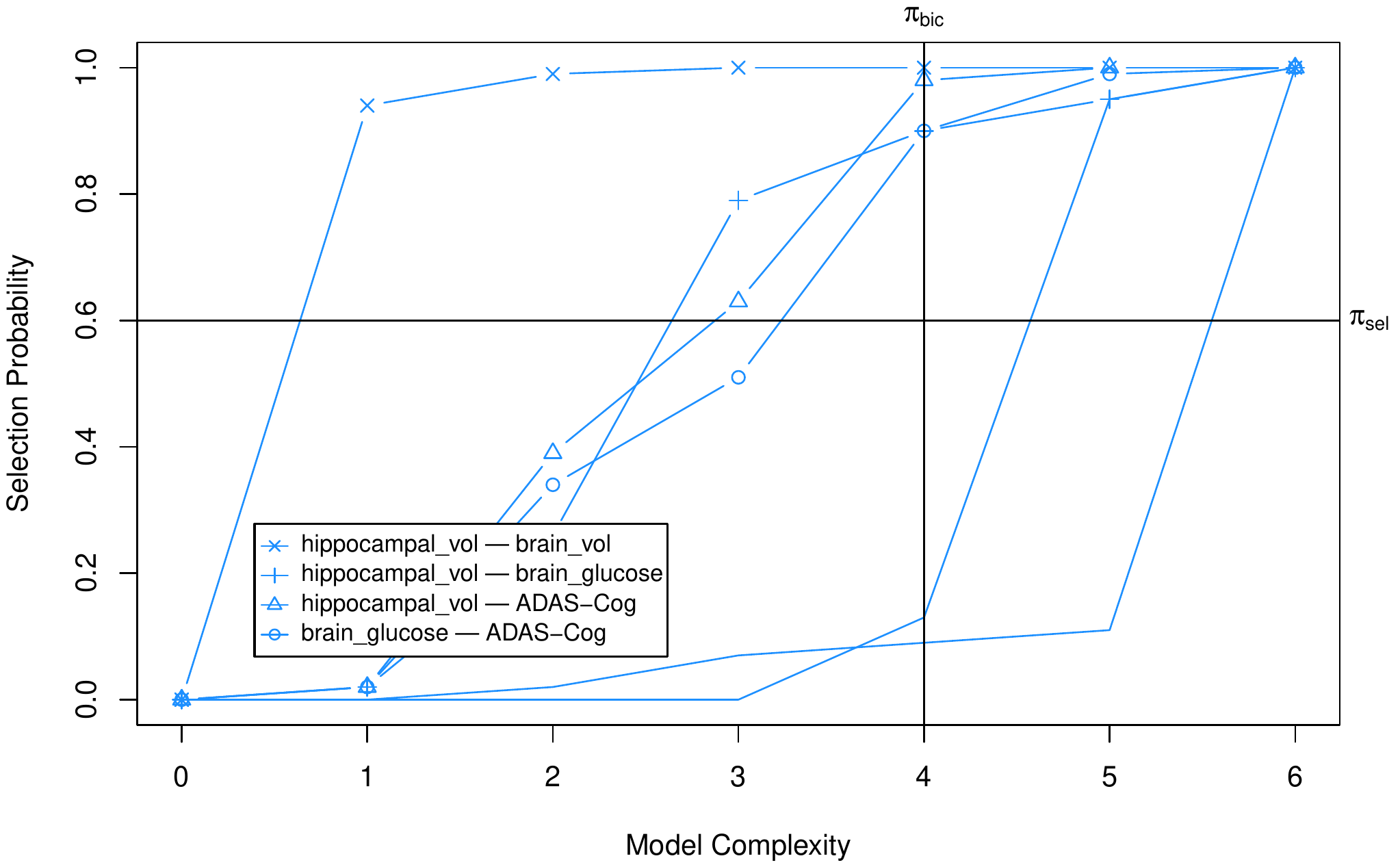}
\label{edgeStabADNIPrior}
}}
\mbox{\subfloat[]
{
\includegraphics[width=0.48\textwidth]{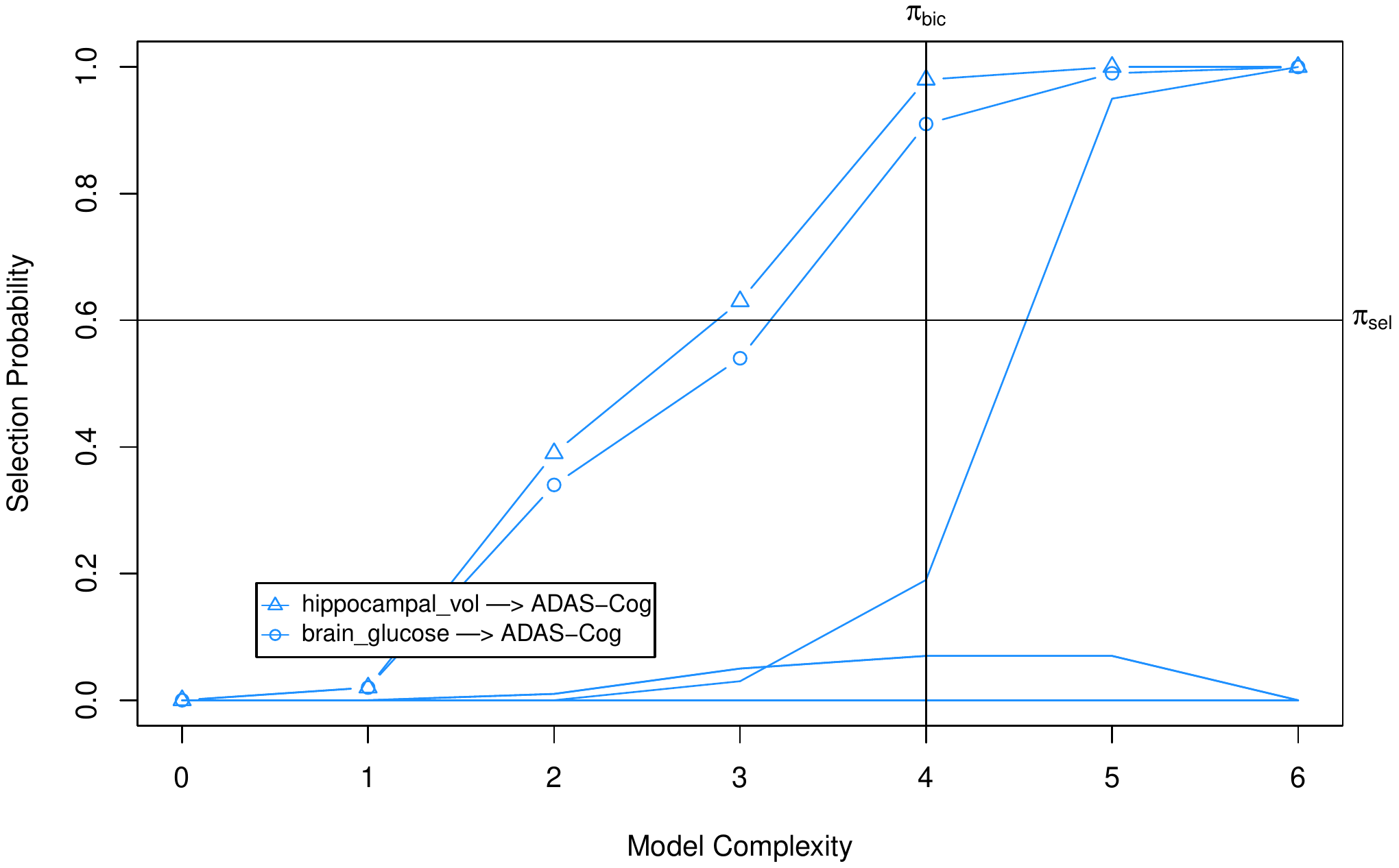}
\label{causalStabADNIPrior}
}}
\mbox{\subfloat[]
{
\includegraphics[width=0.48\textwidth]{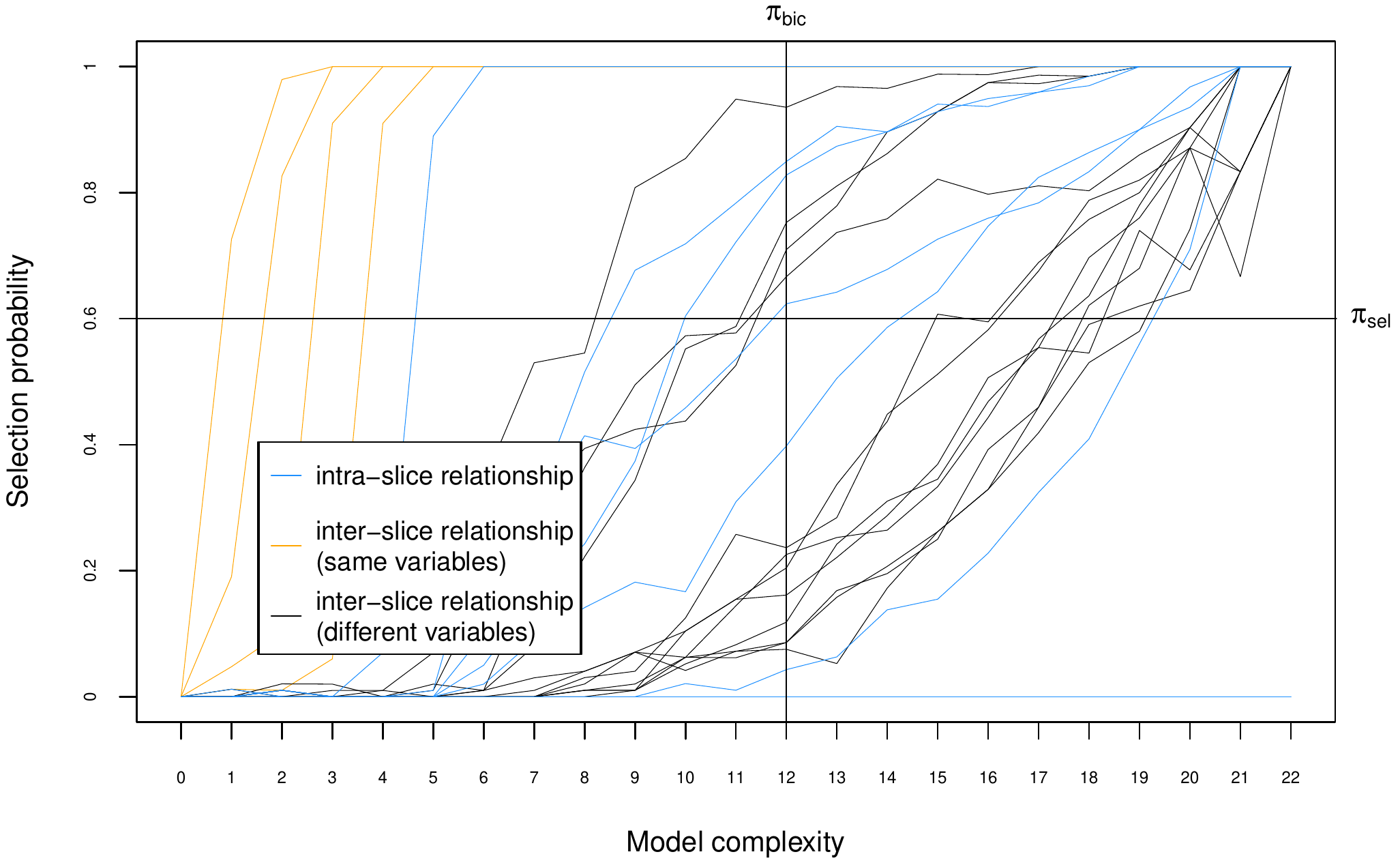}
\label{edgeStabADNITransition}
}}
\mbox{\subfloat[]
{
\includegraphics[width=0.48\textwidth]{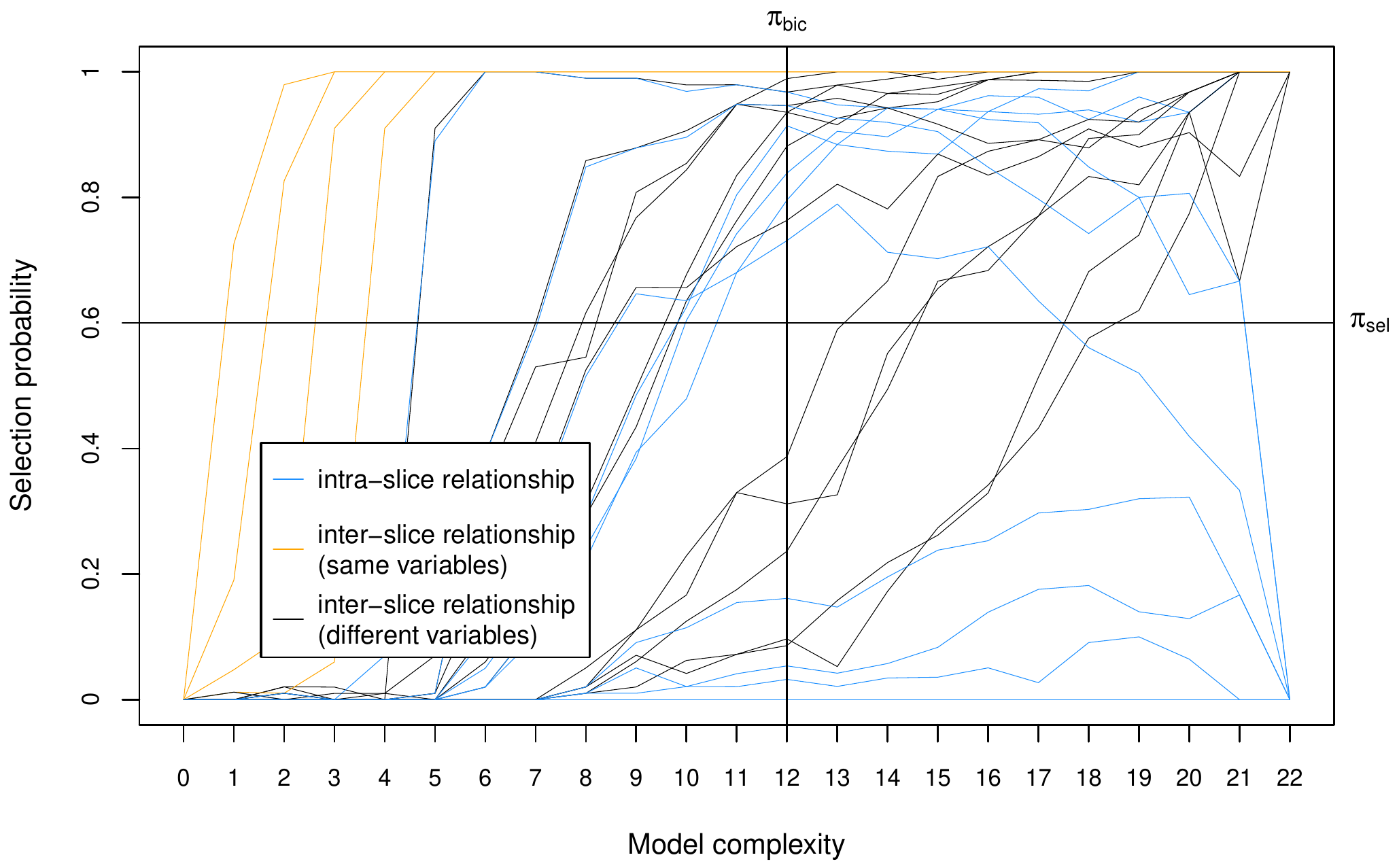}
\label{causalStabADNITransition}
}}
\caption{The stability graphs of the baseline model in (a) and (b) and the transition model in (c) and (d) for Alzheimer's disease, with edge stability in (a) and (c), and causal path stability in (b) and (d). The relevant regions, above $\pi_{\mathrm{sel}}$ and left of $\pi_{\mathrm{bic}}$, contain the relevant structures.
}
\label{stabGraphAD}
\end{figure*}
\begin{figure*}[!htbp]
\centering
\mbox{\subfloat[]
{
\includegraphics[width=0.24\textwidth]{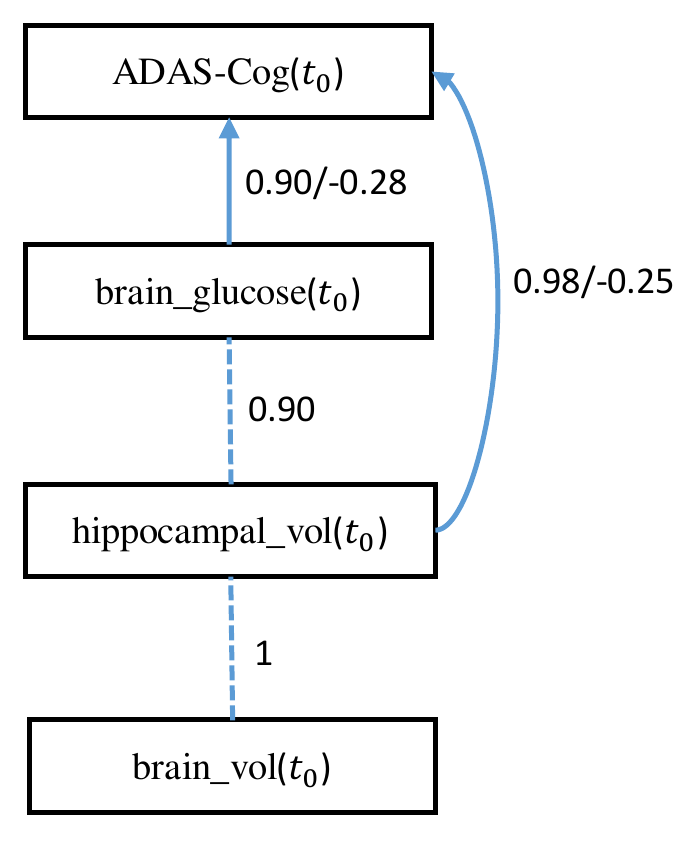}
\label{priorModelAD}
}}
\qquad
\mbox{\subfloat[]
{
\includegraphics[width=0.625\textwidth]{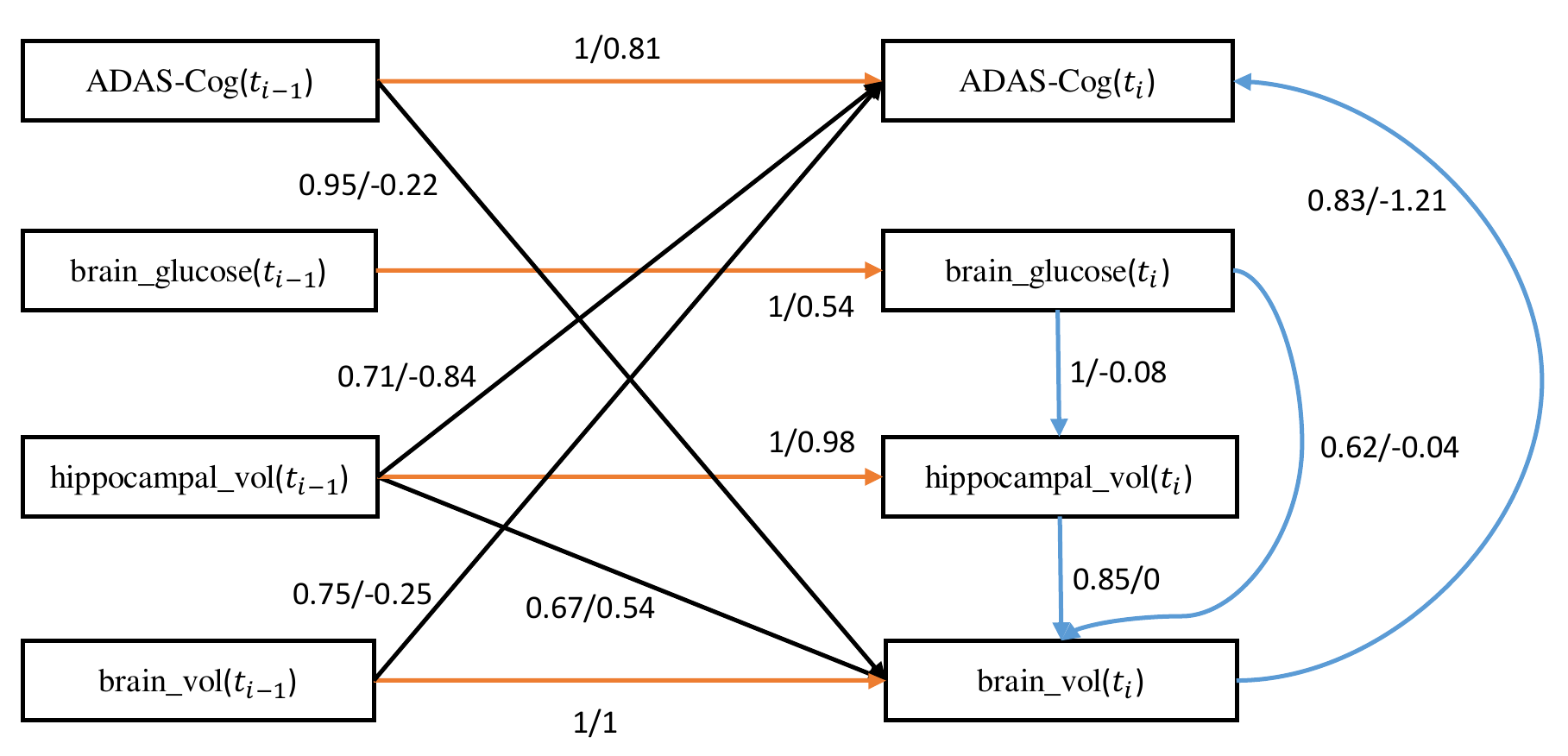}
\label{transitionModelAD}
}}
\caption{(a) The baseline model and (b) the transition model of Alzheimer's disease. The dashed line represents a strong relation between two variables but the causal direction cannot be determined from the data.
Each edge has a reliability score (the highest selection probability in the relevant region of the edge stability graph) and a standardized total causal effect estimation. For example, the annotation \quotes{$1/0.81$} represents a reliability score of $1$ and a total standardized causal effect of $0.81$. Note that the standardized total causal effect represents not just the direct causal effect corresponding to the edge, but the total causal effect also including indirect effects}.
\label{causalModelAD}
\end{figure*}

First we discuss the baseline model which only considers the baseline causal relationships.
The corresponding stability graphs are shown in Figures~\ref{edgeStabADNIPrior} and \ref{causalStabADNIPrior}. $\pi_{\mathrm{sel}}$ is set to $0.6$ and the search phase of S3L found that $\pi_{\mathrm{bic}}=4$. Figures~\ref{edgeStabADNIPrior} and \ref{causalStabADNIPrior} show that four relevant edges and two relevant causal paths were found. Following the visualization procedure, we obtain the baseline model in Figure~\ref{priorModelAD}.
We found that an increase in both brain glucose metabolism and hippocampal volume causes reduction in subject's cognitive dysfunction. These causal relations are consistent with findings in \cite{haight2012relative} which also concluded that both \emph{brain\_glucose} and \emph{hippocampal\_vol} were independently related to \emph{ADAS-Cog} (in our model, it is represented by independent direct causal paths). Additionally, strong relations between hippocampal volume and brain volume seem plausible as they both measure the volume of the brain (partly and entirely).

Next we discuss the transition model which considers all causal relationships across time slices.
We set $\pi_{\mathrm{sel}}=0.6$ and the search phase of S3L yielded $\pi_{\mathrm{bic}}=12$. The corresponding stability graphs can be seen in Figures~\ref{edgeStabADNITransition} and \ref{causalStabADNITransition}. We found twelve relevant edges (see Figure~\ref{edgeStabADNITransition}), consisting of four intra-slice (blue lines)
and eight inter-slice relationships of which four are between the same variables (orange lines) and four are between different variables (black lines). Moreover, we found seventeen relevant causal paths (see Figure~\ref{causalStabADNITransition}), consisting of six intra-slice (blue lines) and eleven inter-slice relationships of which four are between the same variables (orange lines) and seven are between different variables (black lines).
Applying the visualization procedure, we obtain the transition model in Figure~\ref{transitionModelAD}. In addition, the direction of the edge from \emph{brain\_glucose} to \emph{brain\_vol} follows because we do not allow cycles in our model.
We found that there are indirect and direct causal relationships from \emph{hippocampal\_vol} and \emph{brain\_vol} at both $t_{i-1}$ and $t_i$ to \emph{ADAS-Cog} at $t_i$. These particular causal relationships support the hypothesis in \cite{haight2012relative} which says that any changes in both hippocampal volume and brain volume will cause short-term effects on a subject's cognitive dysfunction, both direct and indirect. In the original paper the authors suggested that the indirect causal relationship is through \emph{brain\_glucose}, but our analysis also discovers a potential indirect effect through \emph{brain\_vol}. Interestingly we found that a change in subject's cognitive dysfunction in a previous time slice $t_{i-1}$ causes a reduction in
brain volume in time slice $t_i$.

\subsubsection{Application to chronic kidney disease}
For the third application to real-world data, we consider a longitudinal data set about chronic kidney disease (CKD), provided by the MASTERPLAN study group\cite{peeters2014nurse}. The MASTERPLAN study was initiated in 2004 as a randomized, controlled trial studying the effect of intensified treatment with the aid of nurse practitioners on cardiovascular and kidney outcome in CKD. This intensified treatment regimen addressed eleven possible risk factors for the progression of CKD simultaneously. The study previously showed that this intensified treatment resulted in fewer patients reaching end stage kidney disease compared to standard treatment\cite{peeters2014nurse}.

Here we focus on the potential causal mediators for the protective effect incurred by the intensified treatment with the aid of nurse practitioners. In other words, we aim to identify which of the treatment targets contributed to the observed overall treatment effect. In the present analysis, we include only variables of interest, being treatment status, either nurse practitioner aided care or standard care, as allocated by the randomization procedure (\emph{treatment}), estimated glomerular filtration rate (\emph{gfr})---a marker for overall kidney function, and a variable indicating informative censoring (\emph{inf\_cens}). Informative censoring occurred when patients reached end stage kidney disease requiring renal replacement therapy, such as dialysis or a kidney transplantation, or when they died. Furthermore, we considered treatment targets that were previously hypothesized to contribute most to the overall treatment effect: systolic blood pressure (\emph{sbp}), LDL-cholesterol (\emph{ldl}) and parathyroid hormone (\emph{pth}) concentrations in blood, and protein excretion via urine (\emph{pcr}). In total, there are $497$ subjects with seven variables (both continuous and discrete) over five time slices. The first time slice contains the baseline observations taken before treatment, and the next time slices are the follow-up observations during treatment. Particularly we set the variable \emph{treatment} only at $t_{i-1}$ as it remains the same over all time slices, and the variable \emph{inf\_cens} only at $t_i$ as it is a consequence of previous treatment.
We further added the prior knowledge that \emph{gfr} at $t_i$ does not directly cause any other variables, and that there are no relations between any variable and \emph{inf\_cens} within $t_i$. Both \emph{gfr} and \emph{inf\_cens} are read-out for CKD progression and are within a time slice always the consequence and never the cause of another variable. However, we relax this prior knowledge at time slice $t_0$ as it is a common assumption that without the treatment, \emph{pth} is a consequence of poor kidney function.
The missing data is $5.2\%$ and
a single imputation with EM was conducted to impute the missing values like in applications to CFS and ADNI data. We performed the search over $100$ subsamples of the original data set.
\begin{figure*}[!htbp]
\centering
\mbox{\subfloat[]
{
\includegraphics[width=0.48\textwidth]{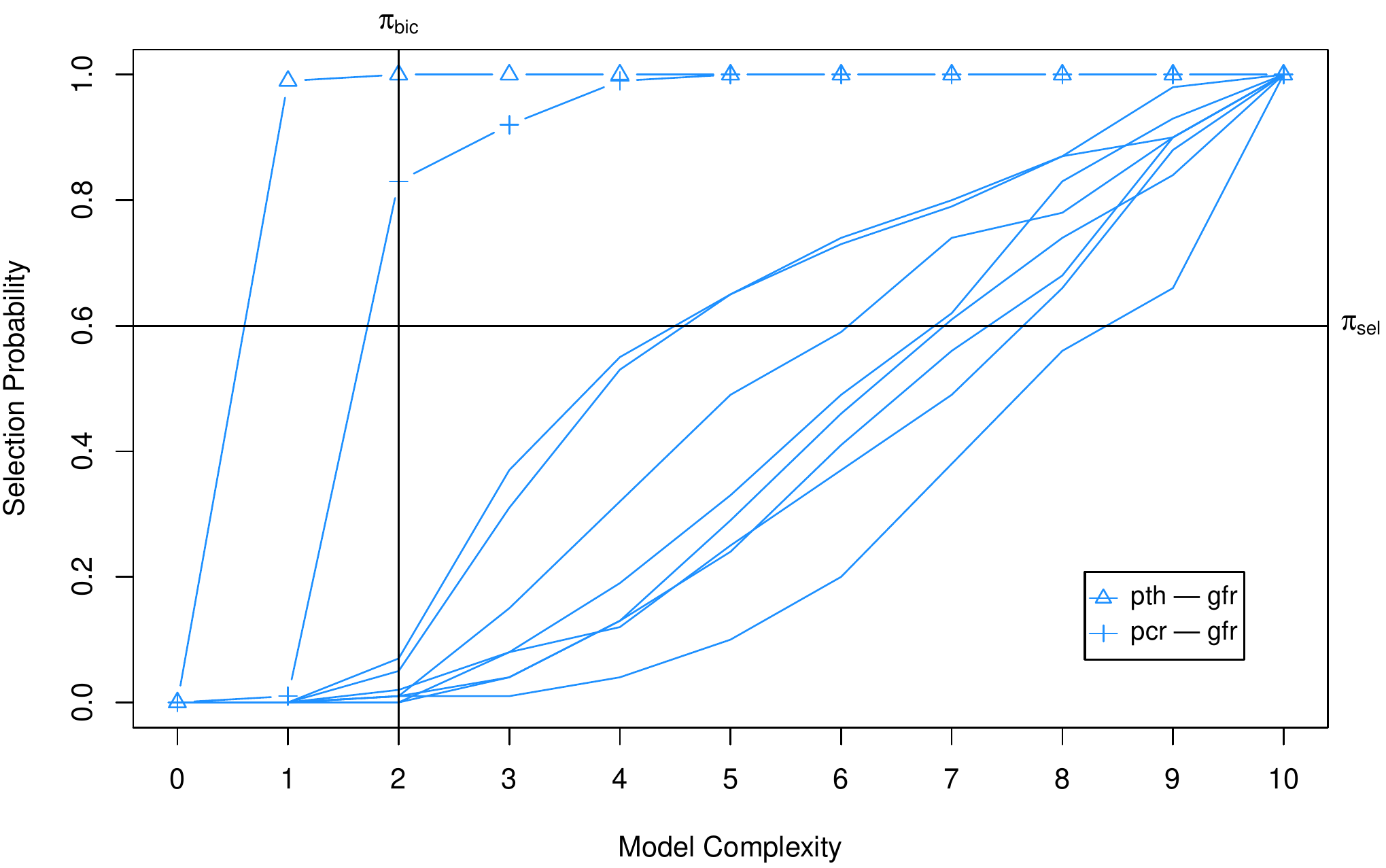}
\label{edgeStabCKDPrior}
}}
\mbox{\subfloat[]
{
\includegraphics[width=0.48\textwidth]{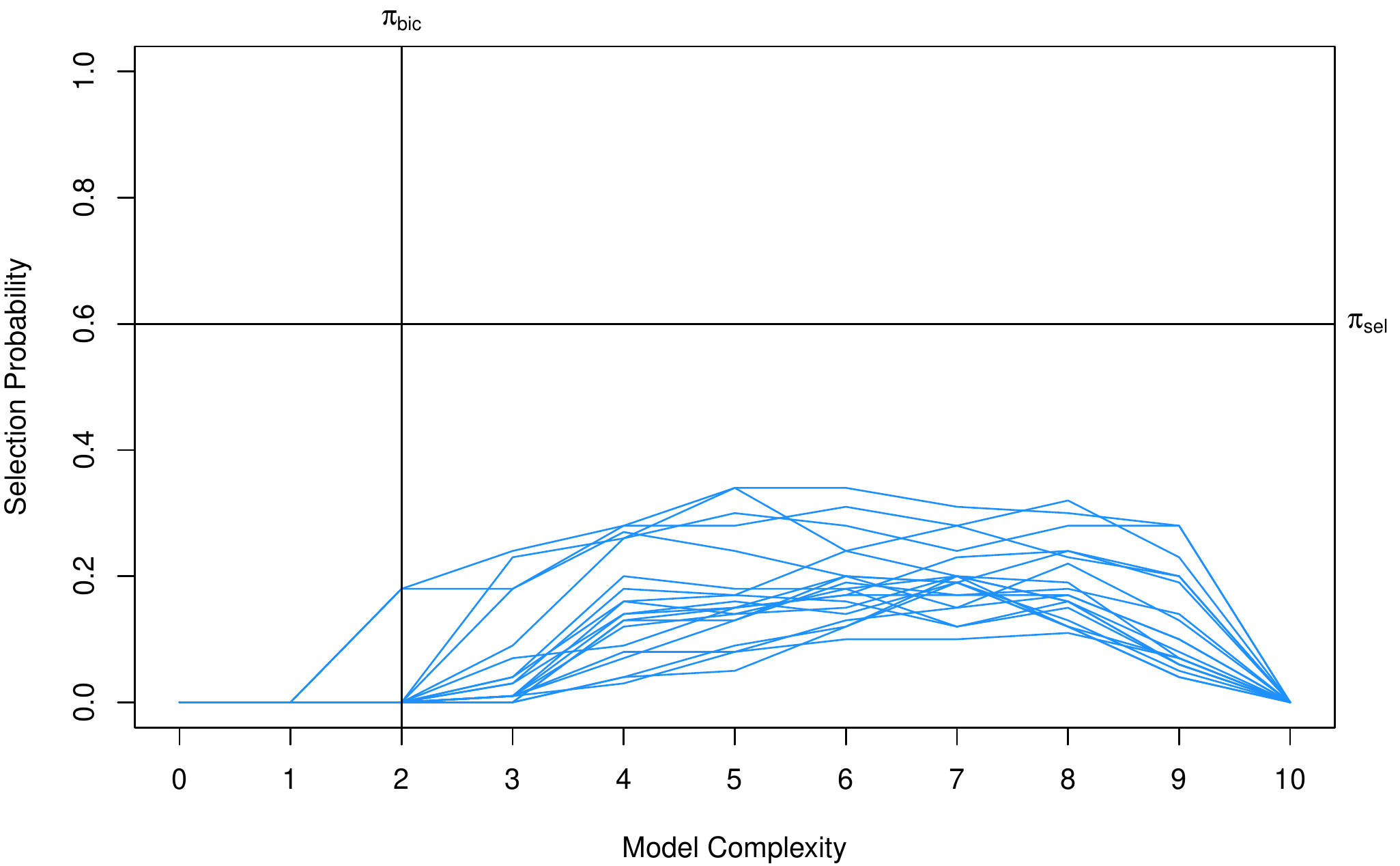}
\label{causalStabCKDPrior}
}}
\mbox{\subfloat[]
{
\includegraphics[width=0.48\textwidth]{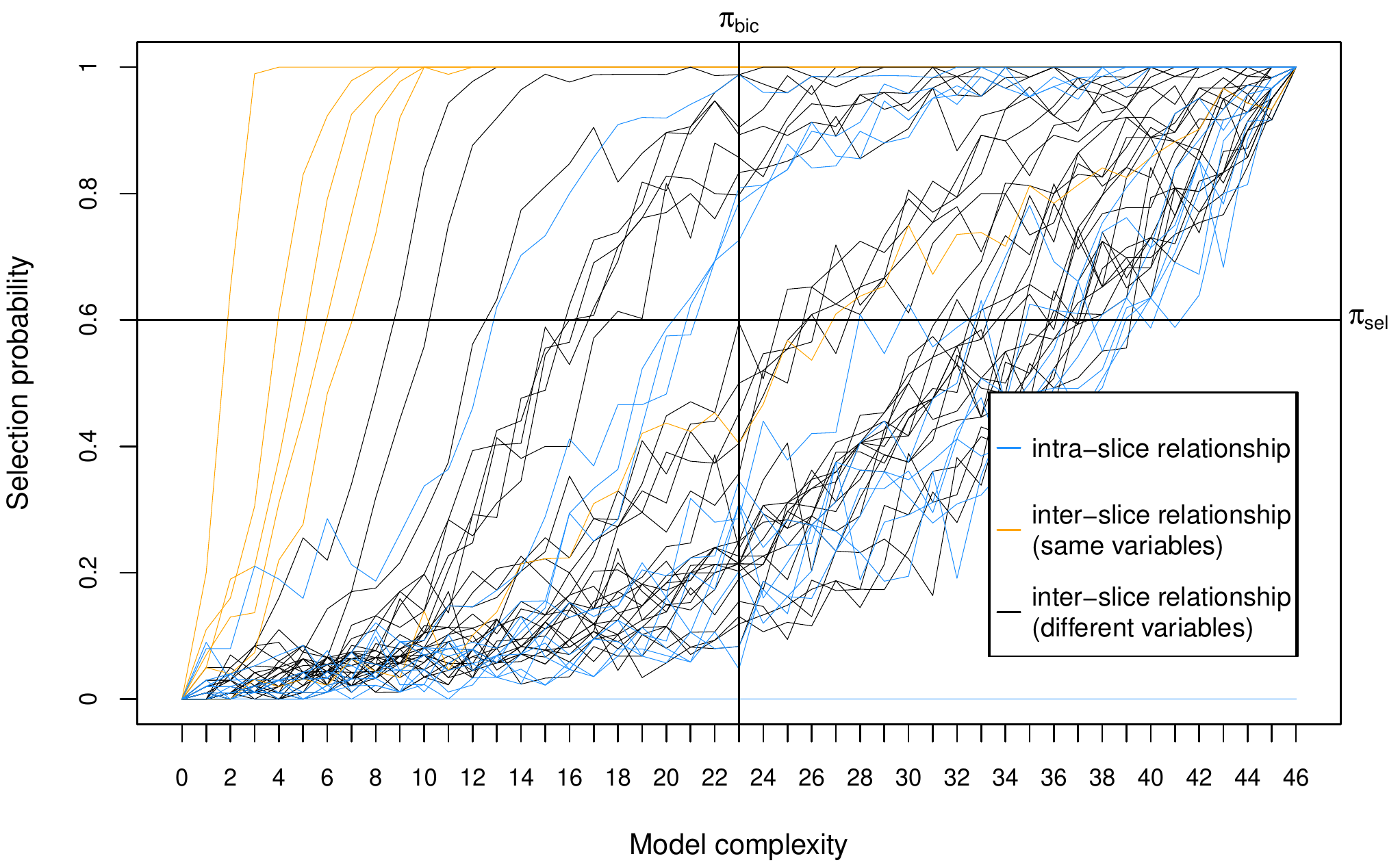}
\label{edgeStabCKDTransition}
}}
\mbox{\subfloat[]
{
\includegraphics[width=0.48\textwidth]{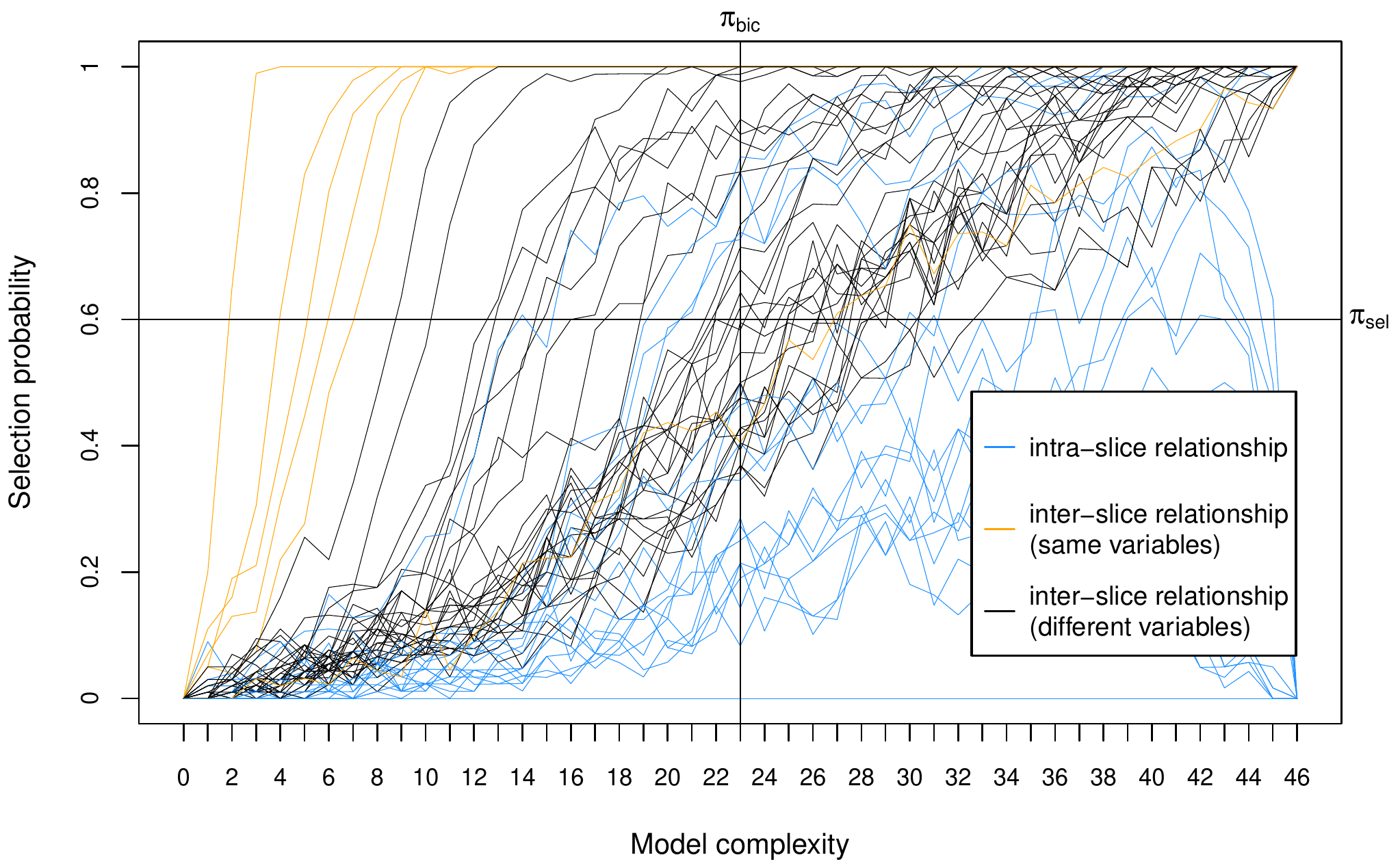}
\label{causalStabCKDTransition}
}}
\caption{The stability graphs of the baseline model in (a) and (b) and the transition model in (c) and (d) for chronic kidney disease, with edge stability in (a) and (c), and causal path stability in (b) and (d). The relevant regions, above $\pi_{\mathrm{sel}}$ and left of $\pi_{\mathrm{bic}}$, contain the relevant structures.
}
\label{stabKidney}
\end{figure*}
\begin{figure*}[!htbp]
\centering
\mbox{\subfloat[]
{
\includegraphics[width=0.22\textwidth]{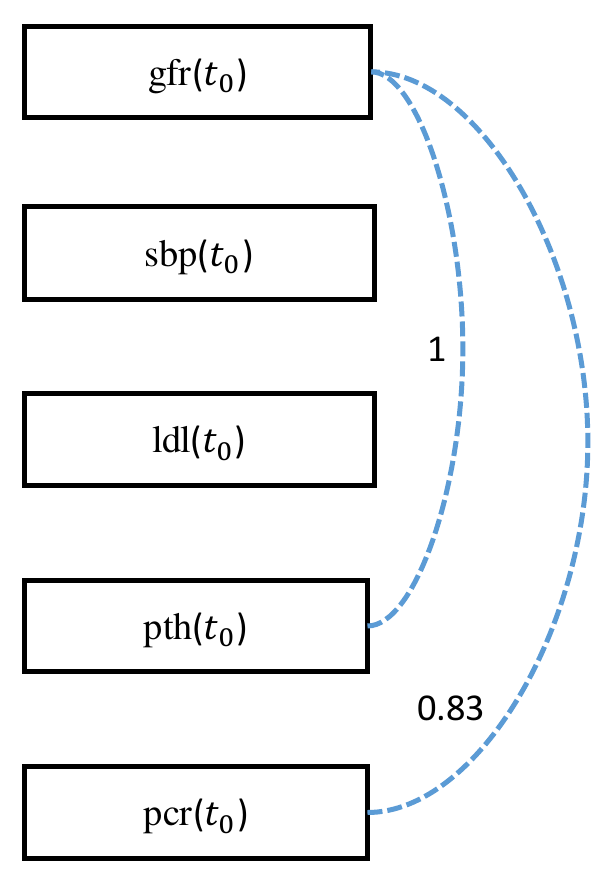}
\label{priorModelCKD}
}}
\qquad
\mbox{\subfloat[]
{
\includegraphics[width=0.6\textwidth]{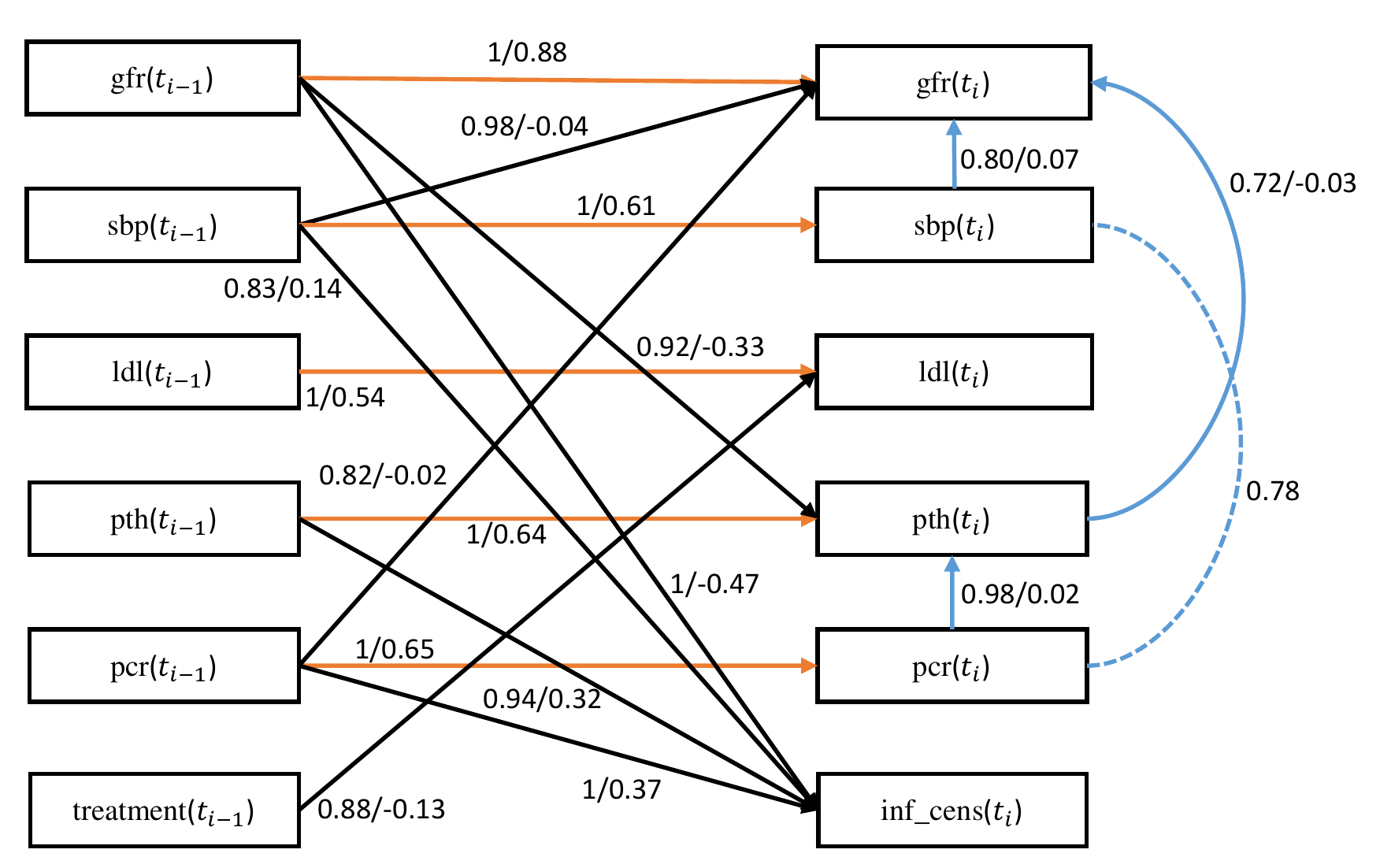}
\label{transitionModelCKD}
}}
\caption{(a) The baseline model and (b) the transition model of chronic kidney disease. The dashed line represents a strong relation between two variables but the causal direction cannot be determined from the data.
Each edge has a reliability score (the highest selection probability in the relevant region of the edge stability graph) and a standardized total causal effect estimation. For example, the annotation \quotes{$1/0.88$} represents a reliability score of $1$ and a standardized total causal effect of $0.88$. Note that the standardized total causal effect represents not just the direct causal effect corresponding to the edge, but the total causal effect also including indirect effects}.
\label{causalModelCKD}
\end{figure*}

First we discuss the baseline model, which only considers the baseline causal relationships. Figures ~\ref{edgeStabCKDPrior} and \ref{causalStabCKDPrior} depict the corresponding stability graphs. As in applications to CFS and ADNI data, $\pi_{sel}$ is set to $0.6$ and based on the search phase of S3L we found that $\pi_{bic}=2$. Figures ~\ref{edgeStabCKDPrior} and \ref{causalStabCKDPrior} shows that two relevant edges were found. Applying the visualization procedure, we get the baseline model in Figure~\ref{priorModelCKD}.
We found that both \emph{pth} and \emph{pcr }were associated with kidney function at baseline. The direction of these associations remains unclear. From renal physiology, we know that proteinuria may result in kidney damage. However, kidney damage and proteinuria may be common consequences of hypertension at an earlier stage in the patient's history. The association between parathyroid hormone and GFR is unsurprising, as calcium and phosphate metabolism is disrupted in patients with advanced kidney disease. However, elevated \emph{pth} may in turn result in further kidney damage by increased vascular calcification. In other words, the associations seem plausible from a physiological point of view, but the association may be in either direction. In the CKD example, a causal direction is almost impossible to ascertain when only using cross-sectional data.

Next we discuss the transition model, which takes into account all causal relationships across time slices. We set $\pi_{\mathrm{sel}}=0.6$ and found $\pi_{\mathrm{bic}}=23$. Based on Figure~\ref{edgeStabCKDTransition}, we obtained seventeen relevant edges, consisting of four intra-slice (blue lines) and thirteen inter-slice relationships of which five are between the same variables (orange lines) and eight are between different variables (black lines). Based on Figure~\ref{causalStabCKDTransition}, we obtained twenty-six relevant causal paths, consisting of five intra-slice (blue lines) and twenty-one inter-slice relationships of which five are between the same variables (orange lines) and sixteen are between different variables (black lines). Applying the visualization procedure, we get the transition model in Figure~\ref{transitionModelCKD}.
Most of the intra-slice and inter-slice causal relationships are very stable with selection probabilities close to $1$. We found inter-slice causal relationships from \emph{gfr}, \emph{sbp}, \emph{pth}, and \emph{pcr} to \emph{inf\_cens}. Furthermore, \emph{gfr}, \emph{sbp}, and \emph{pcr} are well known determinants for CKD progression. The causal relationship from \emph{pth} to \emph{inf\_cens} was somewhat surprising. However, \emph{pth} is a marker for regulation of phosphate stores in the body and related to overall vascular damage through vascular calcification, and may thereby be related to mortality. Indeed, literature indicates that lowering \emph{pth} in dialysis patients resulted in a reduction in mortality\cite{chertow2012effect}. The same may hold true for patients who have CKD and who do yet need dialysis treatment. Perhaps most surprising are the relations between \emph{sbp} and \emph{pcr} and \emph{gfr}, respectively. From renal physiology we know that higher filtration pressures due to higher blood pressure causes the short term glomerular filtration rate to increase slightly\cite{johnson2014comprehensive}. Likewise, at higher filtration pressure, more and larger proteins are pushed out of the blood stream and into the pro-urine and are ultimately excreted via the urine. In the long term, chronically elevated filtration pressures and elevated levels of protein in the pro-urine cause kidney damage and ultimately even end stage kidney disease. Overall, the results are consistent with literature and physiology\cite{levin2013kidney}.

\section{Conclusion and future work}
\label{sec:conclusion}
Causal discovery from longitudinal data turns out to be an important problem in many disciplines. In the medical domain, revealing causal relationships from a given data set may lead to improvement of clinical practice, e.g., further development of treatment and medication. In the past decades, many causal discovery algorithms have been introduced. These causal discovery algorithms, however, have difficulty dealing with the inherent instability in structure estimation.

The present work introduces S3L, a novel discovery algorithm for longitudinal data that is robust for finite samples, extending our previous method\cite{Rahmadi2016} on cross-sectional data.
S3L adopts the concept of stability selection to improve the robustness of structure learning by taking into account a whole range of model complexities. Since finding the optimal model structure for each model complexity is a hard optimization problem, we rephrase stability selection as a multi-objective optimization problem, so that we can jointly optimize over the whole range of model complexities and find the corresponding optimal structures. Moreover, S3L is a general framework that can be combined with alternative approaches, without modifying their original assumptions, e.g., linearity, non-Gaussian noise, etc.

The comparison on the simulated data shows that S3L achieves at least comparable performance as, but often a significant improvement over alternative approaches, mainly in obtaining the causal relations, and in the case of small sample size.
Moreover, the results of experiments on three real-world data sets are corroborated by literature studies\cite{vercoulen1998persistence,wiborg2012towards,heins2013process,haight2012relative,levin2013kidney,henneman2009hippocampal,mungas2002volumetric,rusinek2003regional,chertow2012effect}.

However, the current method considers only longitudinal data with observed variables and cannot handle missing values (other than through imputation as a preprocessing step). We also still assume that the time intervals between time slices is fairly uniform between subjects. Some existing approaches called \emph{random-coefficient} models, also termed \emph{multi-level} or \emph{hierarchical regression models}\cite{raudenbush2002hierarchical,kreft1998introducing}, are flexible to handle unequal intervals between time slices within a subject and/or across subjects.
Future research will aim to account for these aforementioned issues.

\ifCLASSOPTIONcompsoc
  \section*{Acknowledgments}
\else
  \section*{Acknowledgment}
\fi
The authors wish to thank Thaddeus J. Haight, Falma Kemalasari, Joseph Ramsey and two anonymous referees for their valuable discussions, comments, and suggestions.

\begin{flushleft}
The OPTIMISTIC Consortium comprises:
\end{flushleft}
\textbf{Partner 1: Radboud University Medical Centre, The Netherlands},
Ms. Shaghayegh Abghari; Dr. Armaz Aschrafi; Mrs. Sacha Bouman; Ms. Yvonne Cornelissen; Dr. Jeffrey Glennon; Dr. Perry Groot; Prof. Arend Heerschap; Ms. Linda Heskamp; Prof. Tom Heskes; Ms. Katarzyna Kapusta; Mrs. Ellen Klerks; Dr. Hans Knoop; Mrs. Daphne Maas; Mr. Kees Okkersen; Dr. Geert Poelmans, Mr. Ridho Rahmadi; Prof. dr. Baziel van Engelen (Chief Investigator and Partner lead); Dr. Marlies van Nimwegen.\\
\textbf{Partner 2: University of Newcastle upon Tyne, UK},
Dr. Grainne Gorman (Partner lead); Ms. Cecilia Jimenez Moreno; Prof. Hanns Lochmüller; Prof. Mike Trenell; Ms. Sandra van Laar; Ms. Libby Wood.\\
\textbf{Partner 3: Ludwig- Maximilians-Universität München, Germany},
Prof. dr. Benedikt Schoser (Partner lead); Dr. Stephan Wenninger; Dr. Angela Schüller.\\
\textbf{Partner 4: Assistance Publique-Hôpitaux de Paris, France},
Mrs Rémie Auguston; Mr. Lignier Baptiste; Dr. Caroline Barau; Prof. Guillaume Bassez (Partner lead); Mrs. Pascale Chevalier; Ms. Florence Couppey; Ms. Stéphanie Delmas; Prof. Jean-François Deux; Mrs. Celine Dogan; Ms. Amira Hamadouche; Dr. Karolina Hankiewicz; Mrs. Laure Lhermet; Ms. Lisa Minier; Mrs. Amandine Rialland; Mr. David Schmitz.\\
\textbf{Partner 5: University of Glasgow, UK},
Prof. Darren G. Monckton (Partner lead); Dr. Sarah A. Cumming; Ms. Berit Adam.\\
\textbf{Partner 6: The University of Dundee, UK},
Prof. Peter Donnan (Partner lead); Mr. Michael Hannah; Dr. Fiona Hogarth; Dr. Roberta Littleford; Dr. Emma McKenzie; Dr. Petra Rauchhaus; Ms. Erna Wilkie; Mrs. Jennifer Williamson.\\
\textbf{Partner 7: Catt-Sci LTD, UK},
Prof. Mike Catt (Partner lead).\\
\textbf{Partner 8: concentris research management gmbh, Germany},
Mrs. Juliane Dittrich; Ms. Ameli Schwalber (Partner lead).\\
\textbf{Partner 9: The University of Aberdeen, UK},
Prof. Shaun Treweek (Partner lead).

%\begin{dci}
The author(s) declared no potential conflicts of interest with respect to the research, authorship, and/or publication of this article.
%\end{dci}

%\begin{funding}
The author(s) disclosed receipt of the following financial support for the research, authorship, and/or publication of this article: This work was supported, in part, by the DGHE of Indonesia as well as by the European Community's Seventh Framework Programme (FP7/2007-2013) under grant agreement n$^{\circ}$ 305697.

The collection and sharing of brain imaging data used in one of the applications to real-world data was funded by the Alzheimer's Disease Neuroimaging Initiative (ADNI) (National Institutes of Health Grant U01 AG024904) and DOD ADNI (Department of Defense award number W81XWH-12-2-0012). ADNI is funded by the National Institute on Aging, the National Institute of Biomedical Imaging and Bioengineering, and through generous contributions from the following: AbbVie, Alzheimer's Association; Alzheimer's Drug Discovery Foundation; Araclon Biotech; BioClinica, Inc.; Biogen; Bristol-Myers Squibb Company; CereSpir, Inc.; Cogstate; Eisai Inc.; Elan Pharmaceuticals, Inc.; Eli Lilly and Company; EuroImmun; F. Hoffmann-La Roche Ltd and its affiliated company Genentech, Inc.; Fujirebio; GE Healthcare; IXICO Ltd.; Janssen Alzheimer Immunotherapy Research \& Development, LLC.; Johnson \& Johnson Pharmaceutical Research \& Development LLC.; Lumosity; Lundbeck; Merck \& Co., Inc.; Meso Scale Diagnostics, LLC.; NeuroRx Research; Neurotrack Technologies; Novartis Pharmaceuticals Corporation; Pfizer Inc.; Piramal Imaging; Servier; Takeda Pharmaceutical Company; and Transition Therapeutics. The Canadian Institutes of Health Research is providing funds to support ADNI clinical sites in Canada. Private sector contributions are facilitated by the Foundation for the National Institutes of Health (www.fnih.org). The grantee organization is the Northern California Institute for Research and Education, and the study is coordinated by the Alzheimer's Disease Cooperative Study at the University of California, San Diego. ADNI data are disseminated by the Laboratory for Neuro Imaging at the University of Southern California.
%\end{funding}

%\clearpage
%\theendnotes
\bibliographystyle{ieeemod}
\bibliography{mybibfile}

% Generated by IEEEtran.bst, version: 1.13 (2008/09/30)
\begin{thebibliography}{10}
\providecommand{\url}[1]{#1}
\csname url@samestyle\endcsname
\providecommand{\newblock}{\relax}
\providecommand{\bibinfo}[2]{#2}
\providecommand{\BIBentrySTDinterwordspacing}{\spaceskip=0pt\relax}
\providecommand{\BIBentryALTinterwordstretchfactor}{4}
\providecommand{\BIBentryALTinterwordspacing}{\spaceskip=\fontdimen2\font plus
\BIBentryALTinterwordstretchfactor\fontdimen3\font minus
  \fontdimen4\font\relax}
\providecommand{\BIBforeignlanguage}[2]{{%
\expandafter\ifx\csname l@#1\endcsname\relax
\typeout{** WARNING: IEEEtran.bst: No hyphenation pattern has been}%
\typeout{** loaded for the language `#1'. Using the pattern for}%
\typeout{** the default language instead.}%
\else
\language=\csname l@#1\endcsname
\fi
#2}}
\providecommand{\BIBdecl}{\relax}
\BIBdecl

\bibitem{daniel2012using}
R.~M. Daniel, M.~G. Kenward, S.~N. Cousens, and B.~L. De~Stavola, ``Using
  causal diagrams to guide analysis in missing data problems,''
  \emph{Statistical methods in medical research}, vol.~21, no.~3, pp. 243--256,
  2012.

\bibitem{hoover2008causality}
K.~D. Hoover, ``Causality in economics and econometrics,'' \emph{The new
  Palgrave dictionary of economics}, vol.~2, 2008.

\bibitem{abu2003government}
S.~Abu-Bader and A.~S. Abu-Qarn, ``Government expenditures, military spending
  and economic growth: causality evidence from egypt, israel, and syria,''
  \emph{Journal of Policy Modeling}, vol.~25, no.~6, pp. 567--583, 2003.

\bibitem{taguri2015causal}
M.~Taguri, J.~Featherstone, and J.~Cheng, ``Causal mediation analysis with
  multiple causally non-ordered mediators,'' \emph{Statistical methods in
  medical research}, p. 0962280215615899, 2015.

\bibitem{pearl1995causal}
J.~Pearl, ``Causal inference from indirect experiments,'' \emph{Artificial
  intelligence in medicine}, vol.~7, no.~6, pp. 561--582, 1995.

\bibitem{detilleux2016bayesian}
J.~Detilleux, J.-Y. Reginster, A.~Chines, and O.~Bruy{\`e}re, ``A bayesian path
  analysis to estimate causal effects of bazedoxifene acetate on incidence of
  vertebral fractures, either directly or through non-linear changes in bone
  mass density,'' \emph{Statistical methods in medical research}, vol.~25,
  no.~1, pp. 400--412, 2016.

\bibitem{spirtes2010introduction}
P.~Spirtes, ``Introduction to causal inference,'' \emph{The Journal of Machine
  Learning Research}, vol.~11, pp. 1643--1662, 2010.

\bibitem{la2014causal}
S.~la~Bastide-van Gemert, R.~P. Stolk, E.~R. van~den Heuvel, and V.~Fidler,
  ``Causal inference algorithms can be useful in life course epidemiology,''
  \emph{Journal of clinical epidemiology}, vol.~67, no.~2, pp. 190--198, 2014.

\bibitem{sokolova2014causal}
E.~Sokolova, P.~Groot, T.~Claassen, and T.~Heskes, ``Causal discovery from
  databases with discrete and continuous variables,'' in \emph{Probabilistic
  Graphical Models}.\hskip 1em plus 0.5em minus 0.4em\relax Springer, 2014, pp.
  442--457.

\bibitem{cooper2015center}
G.~F. Cooper, I.~Bahar, M.~J. Becich, P.~V. Benos, J.~Berg, J.~U. Espino,
  C.~Glymour, R.~C. Jacobson, M.~Kienholz, A.~V. Lee \emph{et~al.}, ``The
  center for causal discovery of biomedical knowledge from big data,''
  \emph{Journal of the American Medical Informatics Association}, p. ocv059,
  2015.

\bibitem{frees2004longitudinal}
E.~W. Frees, \emph{Longitudinal and panel data: analysis and applications in
  the social sciences}.\hskip 1em plus 0.5em minus 0.4em\relax Cambridge
  University Press, 2004.

\bibitem{fitzmaurice2012applied}
G.~M. Fitzmaurice, N.~M. Laird, and J.~H. Ware, \emph{Applied longitudinal
  analysis}.\hskip 1em plus 0.5em minus 0.4em\relax John Wiley \& Sons, 2012,
  vol. 998.

\bibitem{hedeker2006longitudinal}
D.~Hedeker and R.~D. Gibbons, \emph{Longitudinal data analysis}.\hskip 1em plus
  0.5em minus 0.4em\relax John Wiley \& Sons, 2006, vol. 451.

\bibitem{swanson1997impulse}
N.~R. Swanson and C.~W. Granger, ``Impulse response functions based on a causal
  approach to residual orthogonalization in vector autoregressions,''
  \emph{Journal of the American Statistical Association}, vol.~92, no. 437, pp.
  357--367, 1997.

\bibitem{bessler2002money}
D.~A. Bessler and S.~Lee, ``Money and prices: Us data 1869--1914 (a study with
  directed graphs),'' \emph{Empirical Economics}, vol.~27, no.~3, pp. 427--446,
  2002.

\bibitem{demiralp2003searching}
S.~Demiralp and K.~D. Hoover, ``Searching for the causal structure of a vector
  autoregression,'' \emph{Oxford Bulletin of Economics and statistics},
  vol.~65, no.~s1, pp. 745--767, 2003.

\bibitem{moneta2008graphical}
A.~Moneta, ``Graphical causal models and vars: an empirical assessment of the
  real business cycles hypothesis,'' \emph{Empirical Economics}, vol.~35,
  no.~2, pp. 275--300, 2008.

\bibitem{kim2007unified}
J.~Kim, W.~Zhu, L.~Chang, P.~M. Bentler, and T.~Ernst, ``Unified structural
  equation modeling approach for the analysis of multisubject, multivariate
  functional mri data,'' \emph{Human Brain Mapping}, vol.~28, no.~2, pp.
  85--93, 2007.

\bibitem{moneta2011causal}
A.~Moneta, N.~Chla{\ss}, D.~Entner, and P.~O. Hoyer, ``Causal search in
  structural vector autoregressive models.'' in \emph{NIPS Mini-Symposium on
  Causality in Time Series}, 2011, pp. 95--114.

\bibitem{peters2013causal}
J.~Peters, D.~Janzing, and B.~Sch{\"o}lkopf, ``Causal inference on time series
  using restricted structural equation models,'' in \emph{Advances in Neural
  Information Processing Systems}, 2013, pp. 154--162.

\bibitem{chu2008search}
T.~Chu and C.~Glymour, ``Search for additive nonlinear time series causal
  models,'' \emph{Journal of Machine Learning Research}, vol.~9, no. May, pp.
  967--991, 2008.

\bibitem{hyvarinen2008causal}
A.~Hyv{\"a}rinen, S.~Shimizu, and P.~O. Hoyer, ``Causal modelling combining
  instantaneous and lagged effects: an identifiable model based on
  non-gaussianity,'' in \emph{Proceedings of the 25th international conference
  on Machine learning}.\hskip 1em plus 0.5em minus 0.4em\relax ACM, 2008, pp.
  424--431.

\bibitem{shimizu2006linear}
S.~Shimizu, P.~O. Hoyer, A.~Hyv{\"a}rinen, and A.~Kerminen, ``A linear
  non-gaussian acyclic model for causal discovery,'' \emph{Journal of Machine
  Learning Research}, vol.~7, no. Oct, pp. 2003--2030, 2006.

\bibitem{spirtes2000causation}
P.~Spirtes, C.~N. Glymour, and R.~Scheines, \emph{Causation, prediction, and
  search}.\hskip 1em plus 0.5em minus 0.4em\relax MIT press, 2000, vol.~81.

\bibitem{Rahmadi2016}
\BIBentryALTinterwordspacing
R.~Rahmadi, P.~Groot, M.~Heins, H.~Knoop, and T.~Heskes, ``Causality on
  cross-sectional data: Stable specification search in constrained structural
  equation modeling,'' \emph{Applied Soft Computing}, 2016. doi:
  http://dx.doi.org/10.1016/j.asoc.2016.10.003. [Online]. Available:
  \url{http://www.sciencedirect.com/science/article/pii/S1568494616305130}
\BIBentrySTDinterwordspacing

\bibitem{meinshausen2010stability}
N.~Meinshausen and P.~B{\"u}hlmann, ``Stability selection,'' \emph{Journal of
  the Royal Statistical Society: Series B (Statistical Methodology)}, vol.~72,
  no.~4, pp. 417--473, 2010.

\bibitem{friedman1998learning}
N.~Friedman, K.~Murphy, and S.~Russell, ``Learning the structure of dynamic
  probabilistic networks,'' in \emph{Proceedings of the Fourteenth conference
  on Uncertainty in artificial intelligence}.\hskip 1em plus 0.5em minus
  0.4em\relax Morgan Kaufmann Publishers Inc., 1998, pp. 139--147.

\bibitem{maathuis2009estimating}
M.~H. Maathuis, M.~Kalisch, P.~B{\"u}hlmann \emph{et~al.}, ``Estimating
  high-dimensional intervention effects from observational data,'' \emph{The
  Annals of Statistics}, vol.~37, no.~6A, pp. 3133--3164, 2009.

\bibitem{stekhoven2012causal}
D.~J. Stekhoven, I.~Moraes, G.~Sveinbj{\"o}rnsson, L.~Hennig, M.~H. Maathuis,
  and P.~B{\"u}hlmann, ``Causal stability ranking,'' \emph{Bioinformatics},
  vol.~28, no.~21, pp. 2819--2823, 2012.

\bibitem{gates2011extended}
K.~M. Gates, P.~C. Molenaar, F.~G. Hillary, and S.~Slobounov, ``Extended
  unified sem approach for modeling event-related fmri data,''
  \emph{NeuroImage}, vol.~54, no.~2, pp. 1151--1158, 2011.

\bibitem{ramsey2010six}
J.~D. Ramsey, S.~J. Hanson, C.~Hanson, Y.~O. Halchenko, R.~A. Poldrack, and
  C.~Glymour, ``Six problems for causal inference from fmri,''
  \emph{neuroimage}, vol.~49, no.~2, pp. 1545--1558, 2010.

\bibitem{colombo2014order}
D.~Colombo and M.~H. Maathuis, ``Order-independent constraint-based causal
  structure learning.'' \emph{The Journal of Machine Learning Research},
  vol.~15, no.~1, pp. 3741--3782, 2014.

\bibitem{ramsey2012adjacency}
J.~Ramsey, J.~Zhang, and P.~L. Spirtes, ``Adjacency-faithfulness and
  conservative causal inference,'' \emph{arXiv preprint arXiv:1206.6843}, 2012.

\bibitem{ramsey2016improving}
J.~Ramsey, ``Improving accuracy and scalability of the pc algorithm by
  maximizing p-value,'' \emph{arXiv preprint arXiv:1610.00378}, 2016.

\bibitem{chickering2002learning}
D.~M. Chickering, ``Learning equivalence classes of {B}ayesian-network
  structures,'' \emph{The Journal of Machine Learning Research}, vol.~2, pp.
  445--498, 2002.

\bibitem{ramseymillion}
J.~Ramsey, M.~Glymour, R.~Sanchez-Romero, and C.~Glymour, ``A million variables
  and more: the fast greedy equivalence search algorithm for learning
  high-dimensional graphical causal models, with an application to functional
  magnetic resonance images,'' \emph{International Journal of Data Science and
  Analytics}, pp. 1--9, 2017.

\bibitem{deb2002fast}
K.~Deb, A.~Pratap, S.~Agarwal, and T.~Meyarivan, ``A fast and elitist
  multiobjective genetic algorithm: {NSGA-II},'' \emph{IEEE Transactions on
  Evolutionary Computation}, vol.~6, no.~2, pp. 182--197, 2002.

\bibitem{pearl2000causality}
J.~Pearl, \emph{Causality: models, reasoning and inference}.\hskip 1em plus
  0.5em minus 0.4em\relax Cambridge Univ Press, 2000.

\bibitem{meek1995causal}
C.~Meek, ``Causal inference and causal explanation with background knowledge,''
  in \emph{Proceedings of the Eleventh conference on Uncertainty in artificial
  intelligence}.\hskip 1em plus 0.5em minus 0.4em\relax Morgan Kaufmann
  Publishers Inc., 1995, pp. 403--410.

\bibitem{pearl2003statistics}
J.~Pearl, ``Statistics and causal inference: A review,'' \emph{Test}, vol.~12,
  no.~2, pp. 281--345, 2003.

\bibitem{drasgow1988polychoric}
F.~Drasgow, ``Polychoric and polyserial correlations,'' \emph{Encyclopedia of
  statistical sciences}, 1988.

\bibitem{grefenstette1986optimization}
J.~J. Grefenstette, ``Optimization of control parameters for genetic
  algorithms,'' \emph{Systems, Man and Cybernetics, IEEE Transactions on},
  vol.~16, no.~1, pp. 122--128, 1986.

\bibitem{miller1995genetic}
B.~L. Miller and D.~E. Goldberg, ``Genetic algorithms, tournament selection,
  and the effects of noise,'' \emph{Complex systems}, vol.~9, no.~3, pp.
  193--212, 1995.

\bibitem{kline2011principles}
R.~Kline, \emph{Principles and Practice of Structural Equation Modeling}, ser.
  Methodology in the social sciences.\hskip 1em plus 0.5em minus 0.4em\relax
  Guilford Press, 2011. ISBN 9781606238769

\bibitem{pcalg2012}
\BIBentryALTinterwordspacing
M.~Kalisch, M.~M\"achler, D.~Colombo, M.~H. Maathuis, and P.~B\"uhlmann,
  ``Causal inference using graphical models with the {R} package {pcalg},''
  \emph{Journal of Statistical Software}, vol.~47, no.~11, pp. 1--26, 2012.
  [Online]. Available: \url{http://www.jstatsoft.org/v47/i11/}
\BIBentrySTDinterwordspacing

\bibitem{rcausal2016}
C.~Wongchokprasitti, \emph{rcausal: R-Causal Library}, 2016, r package version
  0.99.8.

\bibitem{fawcett2004roc}
T.~Fawcett, ``{ROC} graphs: Notes and practical considerations for
  researchers,'' \emph{Machine learning}, vol.~31, pp. 1--38, 2004.

\bibitem{delong1988comparing}
E.~R. DeLong, D.~M. DeLong, and D.~L. Clarke-Pearson, ``Comparing the areas
  under two or more correlated receiver operating characteristic curves: a
  nonparametric approach,'' \emph{Biometrics}, pp. 837--845, 1988.

\bibitem{robin2011proc}
X.~Robin, N.~Turck, A.~Hainard, N.~Tiberti, F.~Lisacek, J.-C. Sanchez, and
  M.~M{\"u}ller, ``{pROC}: an open-source package for {R} and {S+} to analyze
  and compare {ROC} curves,'' \emph{BMC bioinformatics}, vol.~12, no.~1, p.~77,
  2011.

\bibitem{venkatraman1996distribution}
E.~Venkatraman and C.~B. Begg, ``A distribution-free procedure for comparing
  receiver operating characteristic curves from a paired experiment,''
  \emph{Biometrika}, vol.~83, no.~4, pp. 835--848, 1996.

\bibitem{fisher1925statistical}
R.~A. Fisher, \emph{Statistical methods for research workers}.\hskip 1em plus
  0.5em minus 0.4em\relax Genesis Publishing Pvt Ltd, 1925.

\bibitem{10.2307/2681650}
\BIBentryALTinterwordspacing
R.~A.~F. Frederick~Mosteller, ``Questions and answers,'' \emph{The American
  Statistician}, vol.~2, no.~5, pp. 30--31, 1948. [Online]. Available:
  \url{http://www.jstor.org/stable/2681650}
\BIBentrySTDinterwordspacing

\bibitem{heins2013process}
M.~J. Heins, H.~Knoop, W.~J. Burk, and G.~Bleijenberg, ``The process of
  cognitive behaviour therapy for chronic fatigue syndrome: Which changes in
  perpetuating cognitions and behaviour are related to a reduction in
  fatigue?'' \emph{Journal of psychosomatic research}, vol.~75, no.~3, pp.
  235--241, 2013.

\bibitem{spss}
\emph{IBM SPSS Statistics for Windows, Version 24.}, IBM Corp., Armonk, NY,
  2016.

\bibitem{vercoulen1998persistence}
J.~Vercoulen, C.~Swanink, J.~Galama, J.~Fennis, P.~Jongen, O.~Hommes,
  J.~Van~der Meer, and G.~Bleijenberg, ``The persistence of fatigue in chronic
  fatigue syndrome and multiple sclerosis: development of a model,''
  \emph{Journal of psychosomatic research}, vol.~45, no.~6, pp. 507--517, 1998.

\bibitem{wiborg2012towards}
J.~F. Wiborg, H.~Knoop, L.~E. Frank, and G.~Bleijenberg, ``Towards an
  evidence-based treatment model for cognitive behavioral interventions
  focusing on chronic fatigue syndrome,'' \emph{Journal of psychosomatic
  research}, vol.~72, no.~5, pp. 399--404, 2012.

\bibitem{weiner2010alzheimer}
M.~W. Weiner, P.~S. Aisen, C.~R. Jack, W.~J. Jagust, J.~Q. Trojanowski,
  L.~Shaw, A.~J. Saykin, J.~C. Morris, N.~Cairns, L.~A. Beckett \emph{et~al.},
  ``The {A}lzheimer's disease neuroimaging initiative: progress report and
  future plans,'' \emph{Alzheimer's \& Dementia}, vol.~6, no.~3, pp. 202--211,
  2010.

\bibitem{petersen1999mild}
R.~C. Petersen, G.~E. Smith, S.~C. Waring, R.~J. Ivnik, E.~G. Tangalos, and
  E.~Kokmen, ``Mild cognitive impairment: clinical characterization and
  outcome,'' \emph{Archives of neurology}, vol.~56, no.~3, pp. 303--308, 1999.

\bibitem{haight2012relative}
T.~J. Haight, W.~J. Jagust, and {Alzheimer's Disease Neuroimaging Initiative},
  ``Relative contributions of biomarkers in {A}lzheimer's disease,''
  \emph{Annals of epidemiology}, vol.~22, no.~12, pp. 868--875, 2012.

\bibitem{peeters2014nurse}
M.~J. Peeters, A.~D. van Zuilen, J.~A. van~den Brand, M.~L. Bots, M.~van Buren,
  M.~A. ten Dam, K.~A. Kaasjager, G.~Ligtenberg, Y.~W. Sijpkens, H.~E. Sluiter
  \emph{et~al.}, ``Nurse practitioner care improves renal outcome in patients
  with {CKD},'' \emph{Journal of the American Society of Nephrology}, vol.~25,
  no.~2, pp. 390--398, 2014.

\bibitem{chertow2012effect}
G.~M. Chertow, G.~A. Block, R.~Correa-Rotter, T.~B. Dr{\"u}eke, J.~Floege,
  W.~G. Goodman, C.~A. Herzog, Y.~Kubo, G.~M. London, K.~W. Mahaffey
  \emph{et~al.}, ``Effect of cinacalcet on cardiovascular disease in patients
  undergoing dialysis.'' \emph{The New England journal of medicine}, vol. 367,
  no.~26, pp. 2482--2494, 2012.

\bibitem{johnson2014comprehensive}
R.~J. Johnson, J.~Feehally, and J.~Floege, \emph{Comprehensive clinical
  nephrology}.\hskip 1em plus 0.5em minus 0.4em\relax Elsevier Health Sciences,
  2014.

\bibitem{levin2013kidney}
A.~Levin, P.~Stevens, R.~Bilous, J.~Coresh, A.~De~Francisco, P.~De~Jong,
  K.~Griffith, B.~Hemmelgarn, K.~Iseki, E.~Lamb \emph{et~al.}, ``{K}idney
  {D}isease: {I}mproving {G}lobal {O}utcomes ({KDIGO}) {CKD} work group: Kdigo
  2012 clinical practice guideline for the evaluation and management of chronic
  kidney disease,'' \emph{Kidney Int Suppl}, vol.~3, no.~1, p. e150, 2013.

\bibitem{henneman2009hippocampal}
W.~Henneman, J.~Sluimer, J.~Barnes, W.~Van Der~Flier, I.~Sluimer, N.~Fox,
  P.~Scheltens, H.~Vrenken, and F.~Barkhof, ``Hippocampal atrophy rates in
  alzheimer disease added value over whole brain volume measures,''
  \emph{Neurology}, vol.~72, no.~11, pp. 999--1007, 2009.

\bibitem{mungas2002volumetric}
D.~Mungas, B.~Reed, W.~Jagust, C.~DeCarli, W.~Mack, J.~Kramer, M.~Weiner,
  N.~Schuff, and H.~Chui, ``Volumetric {MRI} predicts rate of cognitive decline
  related to {AD} and cerebrovascular disease,'' \emph{Neurology}, vol.~59,
  no.~6, pp. 867--873, 2002.

\bibitem{rusinek2003regional}
H.~Rusinek, S.~De~Santi, D.~Frid, W.-H. Tsui, C.~Y. Tarshish, A.~Convit, and
  M.~J. de~Leon, ``Regional brain atrophy rate predicts future cognitive
  decline: 6-year longitudinal mr imaging study of normal aging 1,''
  \emph{Radiology}, vol. 229, no.~3, pp. 691--696, 2003.

\bibitem{raudenbush2002hierarchical}
S.~W. Raudenbush and A.~S. Bryk, \emph{Hierarchical linear models: Applications
  and data analysis methods}.\hskip 1em plus 0.5em minus 0.4em\relax Sage,
  2002, vol.~1.

\bibitem{kreft1998introducing}
I.~G. Kreft and J.~de~Leeuw, \emph{Introducing multilevel modeling}.\hskip 1em
  plus 0.5em minus 0.4em\relax Sage, 1998.

\end{thebibliography}

\end{document}